\documentclass[10pt,twocolumn,letterpaper]{article}

\usepackage{cvpr}              
\usepackage{times}
\usepackage{bm}
\usepackage{microtype}
\usepackage{epsfig}
\usepackage{caption}
\usepackage{float}
\usepackage{placeins}
\usepackage{color, colortbl}
\usepackage{stfloats}
\usepackage{enumitem}
\usepackage{tabularx}
\usepackage{xstring}
\usepackage{multirow}
\usepackage{xspace}
\usepackage{url}
\usepackage[hang,flushmargin]{footmisc}
\usepackage{algorithmic}
\usepackage{algorithm}
\usepackage{footnote}
\usepackage[dvipsnames]{xcolor}

\usepackage[accsupp]{axessibility}
\definecolor{cvprblue}{rgb}{0.21,0.49,0.74}
\usepackage[pagebackref,breaklinks,colorlinks,citecolor=cvprblue]{hyperref}

\def\paperID{6379} 
\def\confName{CVPR}
\def\confYear{2024}

\title{High-fidelity Person-centric Subject-to-Image Synthesis}

\author{
	\fontsize{12}{14}\selectfont
	\textbf{Yibin Wang}\textsuperscript{1, 3}\thanks{Equal contribution.},
	\textbf{Weizhong Zhang}\textsuperscript{2}\footnotemark[1],
	\textbf{Jianwei Zheng}\textsuperscript{3}\footnotemark[2],
	\textbf{Cheng Jin}\textsuperscript{1, 4}\thanks{Corresponding author} \\
	\textsuperscript{1}School of Computer Science, Fudan University 
        \textsuperscript{2}School of Data Science, Fudan University \\
	\textsuperscript{3}College of Computer Science and Technology, Zhejiang University of Technology 
        \textsuperscript{4}Haina Lab \\
	{\tt\small yibinwang1121@163.com, weizhongzhang@fudan.edu.cn, zjw@zjut.edu.cn, jc@fudan.edu.cn}
}

\begin{document}

\twocolumn[{%
	\renewcommand\twocolumn[1][]{#1}%
	\maketitle
	\begin{center}
		\centering
		\captionsetup{type=figure}
		\includegraphics[width=0.8\linewidth]{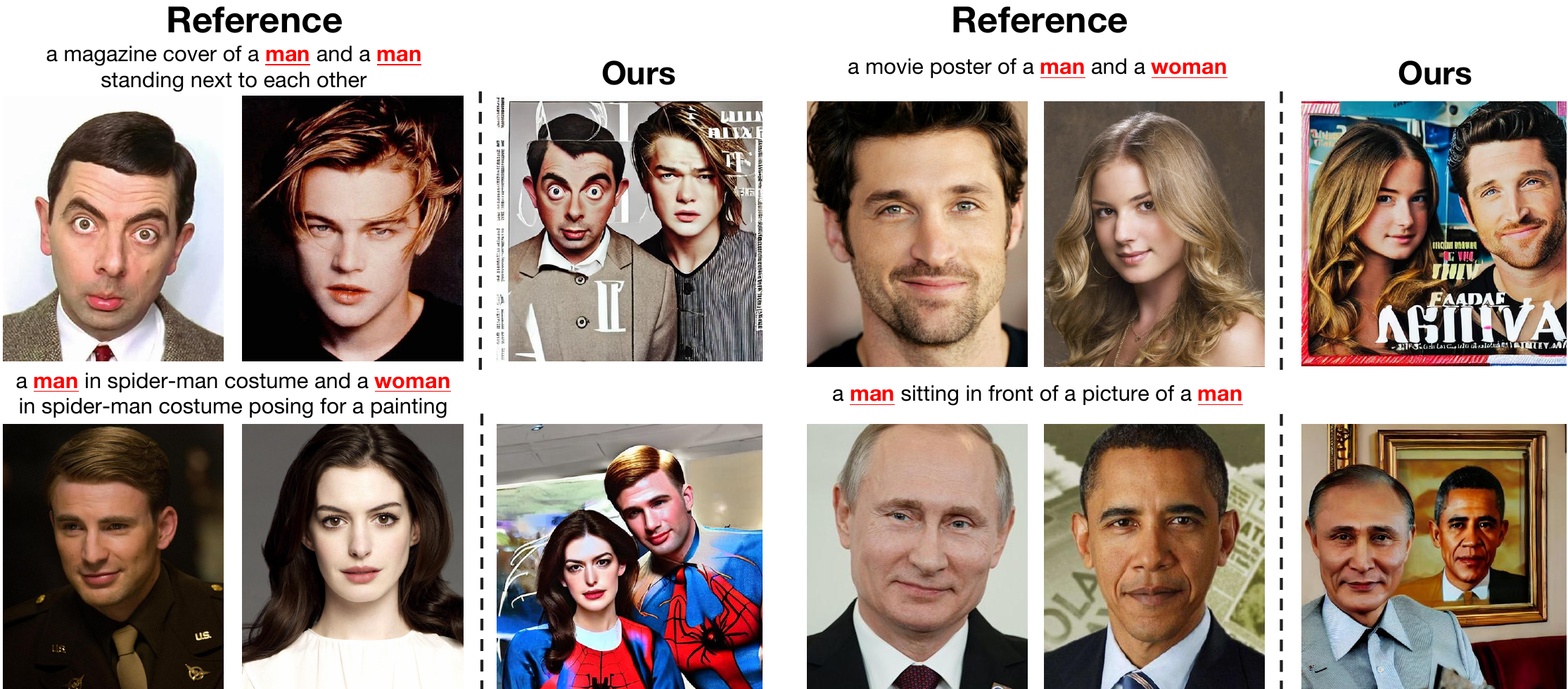}
		\captionof{figure}{Displayed are the results generated using our Face-diffuser, showcasing its prowess across varied inputs. Each instance comprises two distinct inputs: a textual description, and reference images.}
		\label{fig:display}
	\end{center}
}]

\let\thefootnote\relax\footnotetext{$^*$Equal contribution. \\ $^\dag$Corresponding author.}

\begin{abstract}
Current subject-driven image generation methods encounter significant challenges in person-centric image generation. The reason is that they learn the semantic scene and person generation by fine-tuning a common pre-trained diffusion, which involves an irreconcilable training imbalance. Precisely,  to generate realistic persons, they need to sufficiently tune the pre-trained model, which inevitably causes the model to forget the rich semantic scene prior and makes scene generation over-fit to the training data. 
Moreover, even with sufficient fine-tuning, these methods can still not generate high-fidelity persons since joint learning of the scene and person generation also lead to quality compromise. In this paper, we propose  Face-diffuser, an effective collaborative generation pipeline to eliminate the above training imbalance and quality compromise. Specifically, we first develop two specialized pre-trained diffusion models, i.e., Text-driven Diffusion Model (TDM) and Subject-augmented Diffusion Model (SDM), for scene and person generation, respectively.
The sampling process is divided into three sequential stages, i.e., semantic scene construction, subject-scene fusion, and subject enhancement. The first and last stages are performed by TDM and SDM respectively. The subject-scene fusion stage, that is the collaboration achieved through a novel and highly effective mechanism, Saliency-adaptive Noise Fusion (SNF). Specifically, it is based on our key observation that there exists a robust link between classifier-free guidance responses and the saliency of generated images. In each time step, SNF leverages the unique strengths of each model and allows for the spatial blending of predicted noises from both models automatically in a saliency-aware manner, all of which can be seamlessly integrated into the DDIM sampling process. Extensive experiments confirm the impressive effectiveness and robustness of the Face-diffuser in generating high-fidelity person images depicting multiple unseen persons with varying contexts. Code is available at \url{https://github.com/CodeGoat24/Face-diffuser}.

\end{abstract}

\begin{figure*}[ht]
    \centering
    \includegraphics[width=0.8\linewidth]{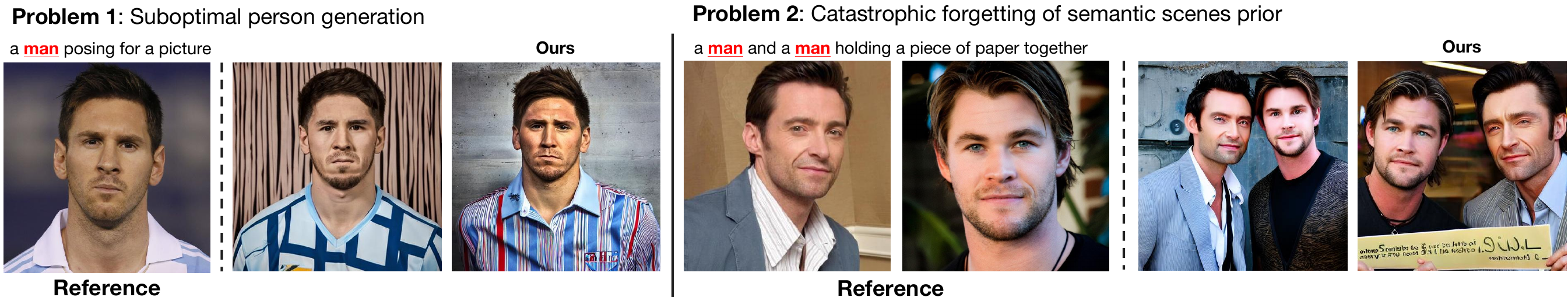}
    \caption{Current methods jointly learn the generation of semantic scenes and persons, which leads to a compromise in the quality of person generation (left), and the irreconcilable training imbalance issue leads to catastrophic forgetting of semantic scenes prior (right). }
    \label{fig:problem}
\end{figure*}

\section{Introduction}
\label{sec:intro}
Subject-driven text-to-image generation \cite{chang2023muse, ding2021cogview, kang2023scaling, sohl2015deep, saharia2022photorealistic, bolya2023token} can synthesize personalized images depicting particular subjects defined by users with a few sample images. The basic idea in optimization-based approaches~\cite{textual_inversion, dreambooth, custom-diffusion, tewel2023key, avrahami2023break, hao2023vico, shi2023instantbooth} is to fine-tune a pre-trained model (e.g., stable diffusion~\cite{ldm}) on a group of provided reference images, typically 3-5 images, for each subject. Another technique roadmap \cite{fastcomposer, wei2023elite, subject-diffusion, chen2023subject, chen2022re} is to retrain a text-to-image generation base model adapted from a pre-trained model with specially-designed structures or retrain the pre-trained model with specific training strategies on a large-scale personalized image dataset. These methods are more practical in real applications since no subject-specific fine-tunings are required in the test time. 

\begin{figure}[t]
    \centering
    \includegraphics[width=1\linewidth]{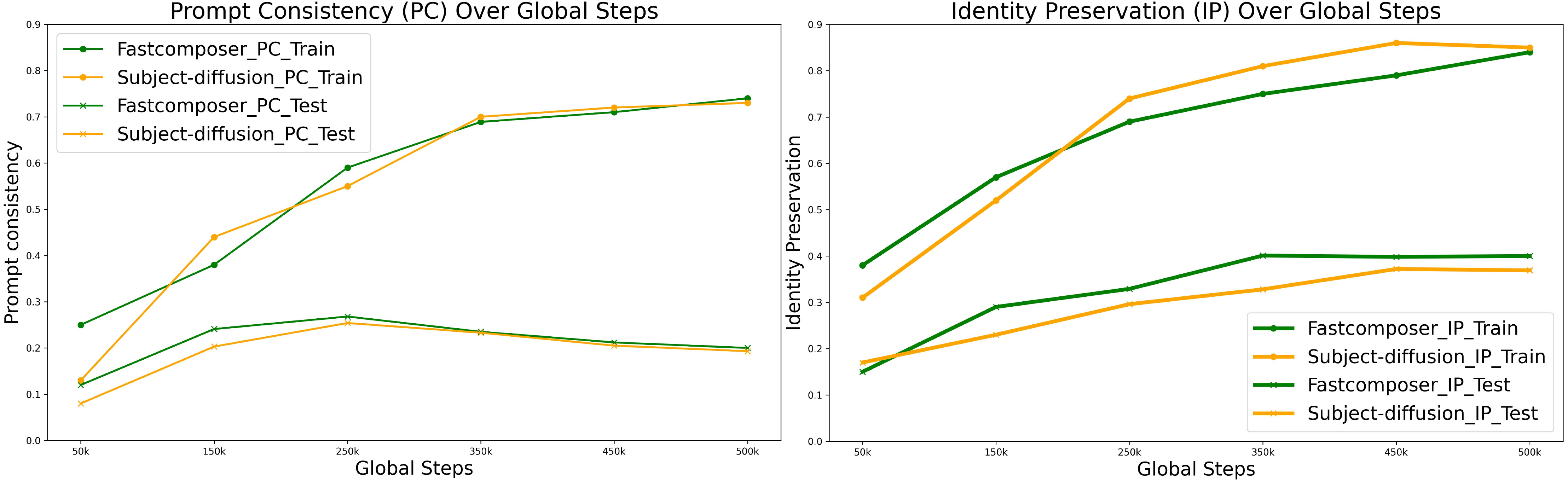}
    \caption{The experimental results showcasing the irreconcilable training imbalance between semantic scene and person generation of Fastcomposer \cite{fastcomposer} and Subject-diffusion \cite{subject-diffusion}. We partitioned the FFHQ-wild \cite{fastcomposer} dataset into training and test sets following a 6:1 ratio and assessed their performance in terms of identity preservation and prompt consistency during continuous training.}
    \label{fig:chart}
\end{figure}

However, we notice that there has been a notable dearth of research on person-centric image generation, and existing subject-driven text-to-image generation models \cite{textual_inversion, custom-diffusion, dreambooth, subject-diffusion, fastcomposer, hao2023vico, wang2023gan} are inadequate for this task. 
To be precise, current state-of-the-art subject-driven text-to-image generation models, e.g., Fastcomposer \cite{fastcomposer} and Subject-diffusion \cite{subject-diffusion}, jointly learn the semantic scene and person generation capability based on some large-scale pre-trained language image model, which has been trained on extensive multimodal datasets like LAION 5B~\cite{schuhmann2022laion}.  But they seem to struggle to effectively harness the inherent prior knowledge of semantic scenes encapsulated within these models. We argue that the reason is these models would over-fit to the textual descriptions and forget the rich semantic scene prior after prolonged training, see \cref{fig:problem} (right) as an example. To validate this hypothesis, we perform experiments to assess their performance in terms of identity preservation and prompt consistency. The results, as illustrated in \cref{fig:chart} (left), show that in the late stage of training, the prompt consistency scores for both models keep increasing for the training data, while exhibiting a noticeable decrease on the test data,  providing strong empirical support for our hypothesis. Obviously, the most straightforward solution appears to be reducing the training duration. Unfortunately, it is proved to be invalid by the irreconcilable training imbalance issue demonstrated in  \cref{fig:chart}. Precisely, \cref{fig:chart} (right) indicates that to generate realistic persons, one needs to
sufficiently tune the pre-trained model, while from \cref{fig:chart} (left), we observe that this inevitably causes the model to forget the rich semantic scene prior and makes scene generation over-fit to the training data. Moreover, even with sufficient fine-tuning, these methods are still unable to generate high-fidelity persons. An example is presented in \cref{fig:problem} (left). The reason could be that jointly learning the generation of semantic scenes and persons may lead to a compromise in the quality of person generation.

To sum up, current methods suffer from catastrophic forgetting of semantic scenes prior due to the irreconcilable training imbalance and the suboptimal person generation due to the quality compromise for joint learning.

In this paper, we propose Face-diffuser, an effective collaborative generation pipeline for person and semantic scene synthesis. To break the training imbalance and quality compromise, we first independently fine-tune two specialized pre-trained diffusion models named Text-driven Diffusion Model (TDM) and Subject-augmented Diffusion Model (SDM) based on stable diffusion \cite{ldm} for scene and person generation, respectively. Recent studies \cite{zheng2023layoutdiffusion, si2023freeu} demonstrate that the generation of an image progresses from the overall scene to intricate details. Following this pipeline, Face-diffuser divides the sampling process into three consecutive stages: semantic scene construction by TDM, subject-scene fusion by collaboration between TDM and SDM, and subject enhancement by SDM. Nevertheless, since SDM and TDM are independent, developing an effective collaboration mechanism for them in the subject-scene fusion stage is the crux of achieving high-fidelity person and diverse semantic scene generation, which is the main contribution of this paper. The details are presented as follows.

It can be expected that to generate high-quality samples, the person and semantic scene generators should collaborate seamlessly in an evolving scheme in test time, that is they should be responsible for different areas in images at different time steps and in generating different images. To achieve this, we propose an effective fine-grained collaborative mechanism named Saliency-adaptive Noise Fusion (SNF) based on classifier-free guidance (CFG), which can be seamlessly integrated into the DDIM sampling process. SNF is motivated by our key observation that the CFG response, i.e., the noise distinction between a given condition and a null condition of each generator can effectively evaluate the impact of the condition on each pixel. Similar finding has also been substantiated in the image editing study~\cite{zhao2023magicfusion}. Therefore, in each step, we generate a saliency-adaptive mask derived from two models' responses of CFG to automatically allocate areas for them to synthesize. Precisely, for each pixel in the image, the responsibility for its synthesis is allocated to the model with the greater response to it in the current step. 

Finally, we would like to highlight that in the test time, to capture and preserve the intricate details of persons given in reference images, we let the CFG response within SDM be the noise distinction between with and without reference person images, thereby neglecting the influence of text condition. This deliberate setting undoubtedly guides SDM to focus the saliency only on person-related areas, resulting in more high-fidelity person generation. Our main contributions can be summarized as follows.

\begin{itemize}
    \item Our proposed Face-diffuser breaks the training imbalance and quality compromise problems in existing subject-to-image generation methods.
    \item We develop two independent models for scene and person generation, and a highly effective collaboration mechanism, Saliency-adaptive Noise Fusion, to utilize each model's strengths for higher-quality image synthesis.
    \item Extensive experiments validate the remarkable effectiveness and robustness of Face-diffuser in generating high-fidelity images portraying multiple unseen persons engaged in diverse contexts.
\end{itemize}

\begin{figure*}[t]
    \centering
    \includegraphics[width=1\linewidth]{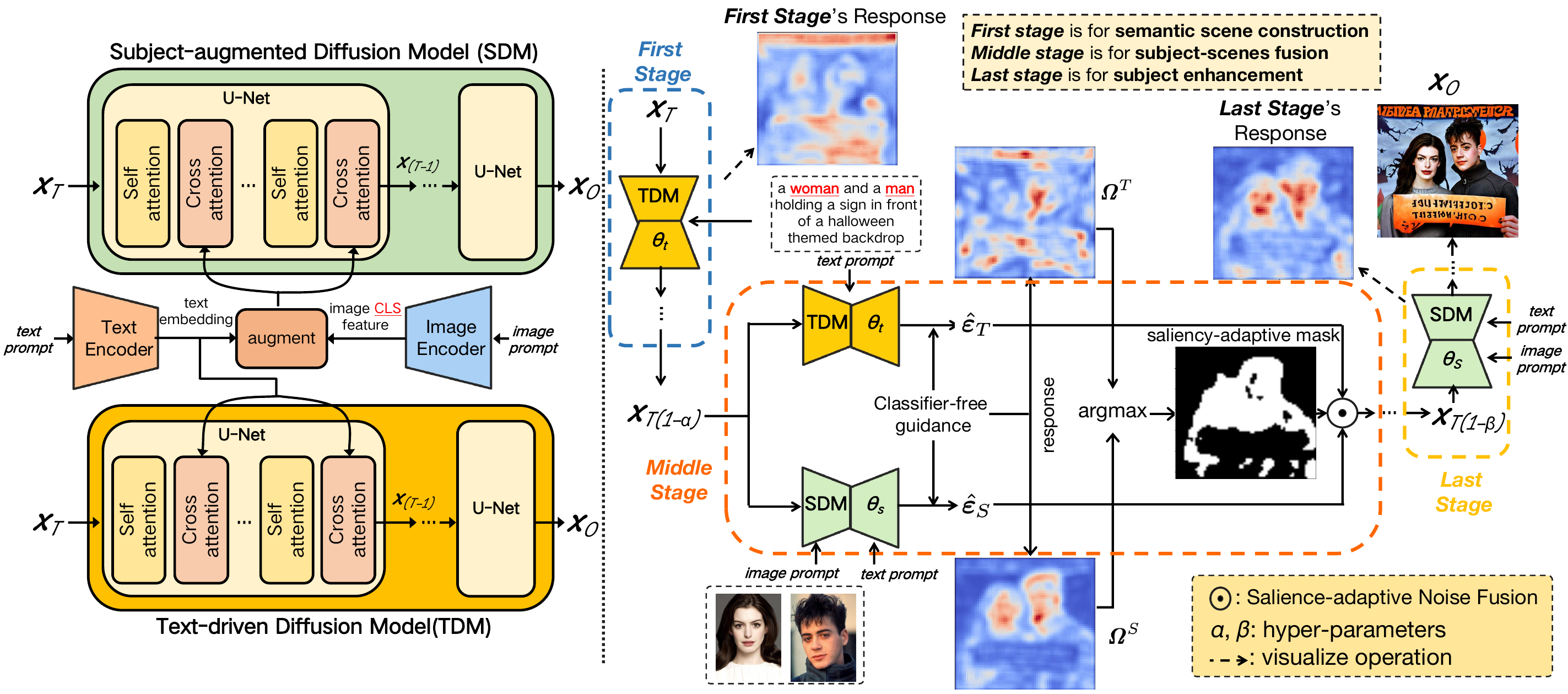}
    \caption{An overview of the Face-diffuser framework. On the left, we display the architectures of two pre-trained models, derived from Stable Diffusion \cite{ldm}, while omitting the autoencoder for simplicity. On the right, we outline our sampling process, which consists of three well-designed stages.}
    \label{fig:model}
   
\end{figure*}

\section{Related Work}
\label{sec:related}
\subsection{Subject-driven Image Generation}
Subject-driven image generation models aim to synthesize personalized images depicting particular subjects defined by users with a few sample images. DreamBooth \cite{dreambooth}, textual-inversion \cite{textual_inversion}, and custom-diffusion \cite{custom-diffusion} employ optimization-based techniques to incorporate subjects into diffusion models. This is achieved through either fine-tuning the model weights \cite{dreambooth, custom-diffusion} or by transforming subject images into a text token encoding the subject's identity \cite{textual_inversion}. However, these models suffer from inefficiency due to the extensive number of fine-tuning steps they demand. In addressing this concern, the tuning-encoder \cite{tuning-encoder} approach mitigates the need for a large number of fine-tuning steps. It achieves this by initially generating an inverted set of latent codes using a pre-trained encoder and subsequently refining these codes through several fine-tuning iterations to better preserve subject identities. However, it's worth noting that all these tuning-based methods \cite{gal2023designing, textual_inversion, custom-diffusion, dreambooth} necessitate resource-intensive backpropagation. This requirement can pose challenges as it demands hardware capable of fine-tuning the model. Such a demand is neither feasible on edge devices, such as smartphones, nor scalable for cloud-based applications. To this end, several concurrent studies have explored tuning-free methods. For instance, X\&Fuse  \cite{kirstain2023x} achieves this by concatenating the reference image with noisy latent variables for image conditioning. ELITE \cite{wei2023elite} and InstantBooth \cite{shi2023instantbooth}, on the other hand, employ global and local mapping networks to project reference images into word embeddings and inject reference image patch features into cross-attention layers to enhance local details. Despite their impressive results in the context of single-object customization, it's important to note that their architectural designs limit their applicability to scenarios involving multiple subjects. This limitation arises due to their reliance on global interactions between the generated image and the reference input image.
In comparison, our Face-Diffuser amortizes the computationally expensive subject tuning during the training phase. This design enables instantaneous personalization for multiple subjects using straightforward feedforward methods during the testing process.

\subsection{Multi-Subject-driven Image Generation}
Custom-diffusion \cite{custom-diffusion} offers the capability of multi-concept composition through joint fine-tuning of the diffusion model for multiple concepts. However, it primarily handles concepts that have clear semantic distinctions, such as animals and their related accessories or backgrounds. On the other hand, SpaText \cite{avrahami2023spatext} and Collage Diffusion \cite{sarukkai2023collage} enable multi-object composition by introducing a layout into the image generation process. These layouts are determined by user-provided segmentation masks and are then transformed into high-resolution images using a diffusion model. Nevertheless, these techniques either compose generic objects with customization \cite{custom-diffusion} or rely on the resource-intensive textual-inversion process to encode instance-specific details \cite{sarukkai2023collage}. Besides, they require users to provide segmentation maps.

Fastcomposer \cite{fastcomposer} offers a solution for generating personalized, multi-subject images in an inference-only manner. It automatically derives plausible layouts from text prompts, eliminating the need for user-provided segmentation maps. Building upon these advancements, Subject-diffusion \cite{subject-diffusion} introduces a method that integrates text and image semantics, incorporating coarse location and fine-grained reference image control to enhance subject fidelity and generalization. However, they face notable challenges in person-centric image generation. Specifically, their models jointly learn the semantic scene and person generation, which leads to training imbalance and quality compromises. In contrast, our Face-diffuser excels at generating high-fidelity characters in diverse semantic scenes without the need for fine-tuning, all in an inference-only manner.

\section{Face-diffuser}
\label{sec:method}
In this section, we will first elaborate on the overview of our architecture as illustrated in \cref{fig:model}. 
After that, we will introduce two specialized models in our pipeline, i.e., Text-driven Diffusion Model (TDM) and Subject-augmented Diffusion Model (SDM). Last, we will delve into the three-stage sampling process, with a focus on the details of our proposed collaboration synthesis mechanism, Saliency-adaptive Noise Fusion (SNF).

\subsection{Overview}
In the training stage, as shown in \cref{fig:model} (left), to eliminate the training imbalance, we {\it independently} fine-tune the pre-trained Stable-Diffusion \cite{ldm} as our TDM  and another similar model with an additional image prompt as our SDM. These two base models will be responsible for the semantic scene and person generation, respectively. 

In the test time (\cref{fig:model} (right)), unlike previous studies \cite{custom-diffusion, dreambooth}, Face-diffuser eliminates the need for subject-specific fine-tuning. To be precise, our sampling process comprised three sequential stages with total  \textit{T} denoising steps.
In the first stage, we employ TDM to construct the preliminary semantic scene for  $\alpha\textit{T}$ steps.

Then, TDM and SDM collaborate to infuse the person into the scene in the following $\beta\textit{T}$ steps based on  our effective collaboration mechanism SNF. At each step, SNF utilizes the responses from classifier-free guidance of both models to produce a saliency-adaptive mask. This mask, with the same size as the target image, serves as a saliency indicator, containing values of 0 and 1 that denote the regions generated by TDM and SDM, respectively.  
They are responsible for different areas at different time steps for flexible evolving collaboration.

In the last stage, SDM is further utilized to refine the quality of generated persons.

\subsection{Semantic Scene and Person Generators}
\subsubsection{Text-driven Diffusion Model (TDM)}
Stable diffusion (SD) is employed as our TDM. For semantic scene generation, given the semantic scene prompt \textit{c} and the input image \textit{\textbf{x}}, the VAE first encodes the \textit{\textbf{x}} into a latent representation \textit{\textbf{z}}, perturbed by Gaussian noise $\bm{\varepsilon}$ to get $\textbf{\textit{z}}_{t}$ at \textit{t} step during diffusion. 
Then the text encoder $\psi$ maps semantic scene prompts \textit{c} to conditional embeddings $\psi(c)$ which would be integrated into the denoiser $\varepsilon_{\theta}$, U-Net through cross-attention \cite{vaswani2017attention, wang2023plsnet, LCANet, DMINet}. The training objective is to minimize the loss function as follows:
\begin{equation}\label{L_noise}
    	L_{noise} = \mathbb{E}_{\textbf{\textit{z}}, c,\bm{\varepsilon} \sim N(0,1), t}  \parallel \bm{\varepsilon} - \varepsilon_{\theta}(\textbf{\textit{z}}_{t}, t, \psi(c)) \parallel_{2}^{2}  \nonumber
\end{equation}
During inference, a random noise $\textbf{\textit{z}}_{T}$ is sampled from a normal distribution \textit{N (0, 1)}, and this noise is iteratively denoised by the U-Net to produce the initial latent representation $\textbf{\textit{z}}_{0}$. Subsequently, the VAE decoder maps these latent codes back to pixel space to generate the final image. 

\subsubsection{Subject-augmented Diffusion Model (SDM)}
The SDM model tailored for subject generation is also based on the SD model but includes an additional reference image condition \textit{r}. Inspired by previous works like \cite{fastcomposer, subject-diffusion}, we adopt a tuning-free approach by enhancing text prompts with visual features extracted from reference images. When given a text prompt and a list of reference images, we begin by encoding the text prompt and reference subjects into embeddings using pre-trained CLIP text and image encoders, respectively. Following this, we replace the user-specific word embeddings with these visual features and input the resulting augmented embeddings into a multilayer perceptron (MLP).  This process yields the final conditioning embeddings, denoted as $\psi(c)_{aug}$. The loss function of SDM closely resembles the one in \cref{L_noise}, with the substitution of $\psi(\textit{c})$ by $\psi(c)_{aug}$.
\begin{equation}\label{L_noise}
    	L_{noise} = \mathbb{E}_{\textbf{\textit{z}}, c,\bm{\varepsilon} \sim N(0,1), t}  \parallel \bm{\varepsilon} - \varepsilon_{\theta}(\textbf{\textit{z}}_{t}, t, \psi(c)_{aug}) \parallel_{2}^{2} 
     \nonumber
\end{equation}

\subsubsection{Condition Effectiveness  Enhancement}
For TDM, to reinforce the effectiveness of the conditions in scene generation, Classifier-free Guidance (CFG) is employed in each step to extrapolate the predicted noise along the direction specified by certain conditions. 
Specifically, at step $t$, CFG takes the form of 
\begin{align}
    \hat{\bm{\varepsilon}}_T={\varepsilon}_{\theta}({\textbf{\textit{z}}}_{t}|\varnothing )+s(\underbrace{{\varepsilon}_{\theta}({\textbf{\textit{z}}}_{t}|c)-{\varepsilon}_{\theta}({\textbf{\textit{z}}}_{t}|\varnothing )}_{
    \textbf{\textit{R}}_T}),  \label{TDM_classifier}
\end{align}
where $\varnothing$ signifies a null condition. The hyperparameter $s>0$ denotes the guidance weight, and the reinforcing effect becomes stronger when $s$ increases.



For SDM, we notice that it is conditioned on both the text and reference images. To eliminate the impact of text condition $c$ and emphasize the noise distinction ablating the reference person images only, we extend CFG as follows:

\begin{align}
    \hat{\bm{\varepsilon}}_{S}={\varepsilon}_{\theta}({\textbf{\textit{x}}}_{t}|\varnothing )+s(\underbrace{{\varepsilon}_{\theta}({\textbf{\textit{x}}}_{t}|c, r)-{\varepsilon}_{\theta}({\textbf{\textit{x}}}_{t}|c )}_{\textbf{\textit{R}}_S}).\label{SDM_classifier} 
\end{align}
This careful design undoubtedly strengthens its ability to generate higher-fidelity characters, as this leads the model to become more adept at capturing and preserving the subtle details of the reference images.

We call the predicted noise differences $\textbf{\textit{R}}_T$ and $\textbf{\textit{R}}_S$ in Eqn.(\ref{TDM_classifier}) and (\ref{SDM_classifier}) the responses of the semantic scene and reference image conditions. They play a fundamental role in developing our collaborative mechanism in Face-Diffuser, which is represented in the next subsection. 

\subsection{Collaboratively Synthesis}
Face-Diffuser employs the following 3 stages to generate each image: 

\textbf{Stage \uppercase\expandafter{\romannumeral1}}: Given initial noise $\textbf{\textit{x}}_{T}$, we first employ TDM to construct the scene for $\alpha\textit{T}$ steps and output $\textbf{\textit{x}}_{T(1-\alpha)}$.

\textbf{Stage \uppercase\expandafter{\romannumeral2}}: After construction of the preliminary scene, we take previous stage's output as input and leverage TDM and SDM to collaboratively infuse the person into the scene for $(\beta-\alpha)T$ steps, outputting $\textbf{\textit{x}}_{T(1-\beta)}$. 

\textbf{Stage \uppercase\expandafter{\romannumeral3}}: Taking $\textbf{\textit{x}}_{T(1-\beta)}$ as input, SDM is further utilized to enhance the details generation of persons in the remaining steps.

In the following, we will delve into the collaboration details between TDM and SDM in the middle stage. 

\subsubsection{Saliency-adptive Noise Fusion}
Note that the responses $\textbf{\textit{R}}_T$ and $\textbf{\textit{R}}_S$ actually evaluate the impact of the semantic scene and reference images on each pixel of the predicted noises, the regions with large values mean the conditions have significant impacts on these pixels, which naturally defines the responsibility of TDM and STM in this step. 

Formally, we first define the  following two salience maps based on $R_T$ and $R_S$:
\begin{align} 
    \bm{\mathit{\Omega}}^{T}\, &=\, \textup{Smooth}(\textup{Abs}(\textbf{\textit{R}}_{T})), \label{omega_T}\\
    \bm{\mathit{\Omega}}^{S}\, &=\, \textup{Smooth}(\textup{Abs}(\textbf{\textit{R}}_{S}))\label{omega_S},
\end{align}
where the operator Abs(·) calculates the absolute values of the input variables, while the Smooth(·) function is applied to reduce high-frequency noise, effectively eliminating local outliers and enhancing the coherence of adjacent regions. The empirical validations of $\bm{\mathit{\Omega}}^T$ and $\bm{\mathit{\Omega}}^S$, i.e., their visualizations are presented in \cref{effectiveness}. 

Given $\bm{\mathit{\Omega}}^T$ and $\bm{\mathit{\Omega}}^S$, we proceed to develop the  saliency-adaptive fusion mask through a comparison between these two salience maps:
\begin{align}
    \textbf{\textit{M}} = \textup{argmax}(\textup{Softmax}(\bm{\mathit{\Omega}}^{T}), \textup{Softmax}(\bm{\mathit{\Omega}}^{S})). \nonumber
\end{align}
The softmax operation here is crucial, as the values of $\bm{\mathit{\Omega}}^T$ and $\bm{\mathit{\Omega}}^S$ could have different magnitudes, it ensures that the sum of each salience map remains constant thus let them comparable.  The mask $\textbf{\textit{M}}$ is adopted to define the collaboration mechanism, i.e,  the pixels with $\textit{M}_{ij} =0$ and $\textit{M}_{ij} =1$ are allocated to TDM and SDM in generation, respectively. Finally, the fused noise can be obtained through the following process:
\begin{align}
    \hat{\bm{\varepsilon}} = \textbf{\textit{M}} \odot \hat{\bm{\varepsilon}}_{S} + (1 - \textbf{\textit{M}}) \odot \hat{\bm{\varepsilon}}_{T},
\end{align}
where $\odot$ denotes Hadamard Product, and we omit \textit{t} for simplicity. It is essential to note that in each sampling step, both models take the blended $\textbf{\textit{z}}_{t}$ as input, facilitating the automatic semantic alignment of the two models' noise space. 

\begin{figure*}[t]
    \centering
    \includegraphics[width=0.8\linewidth]{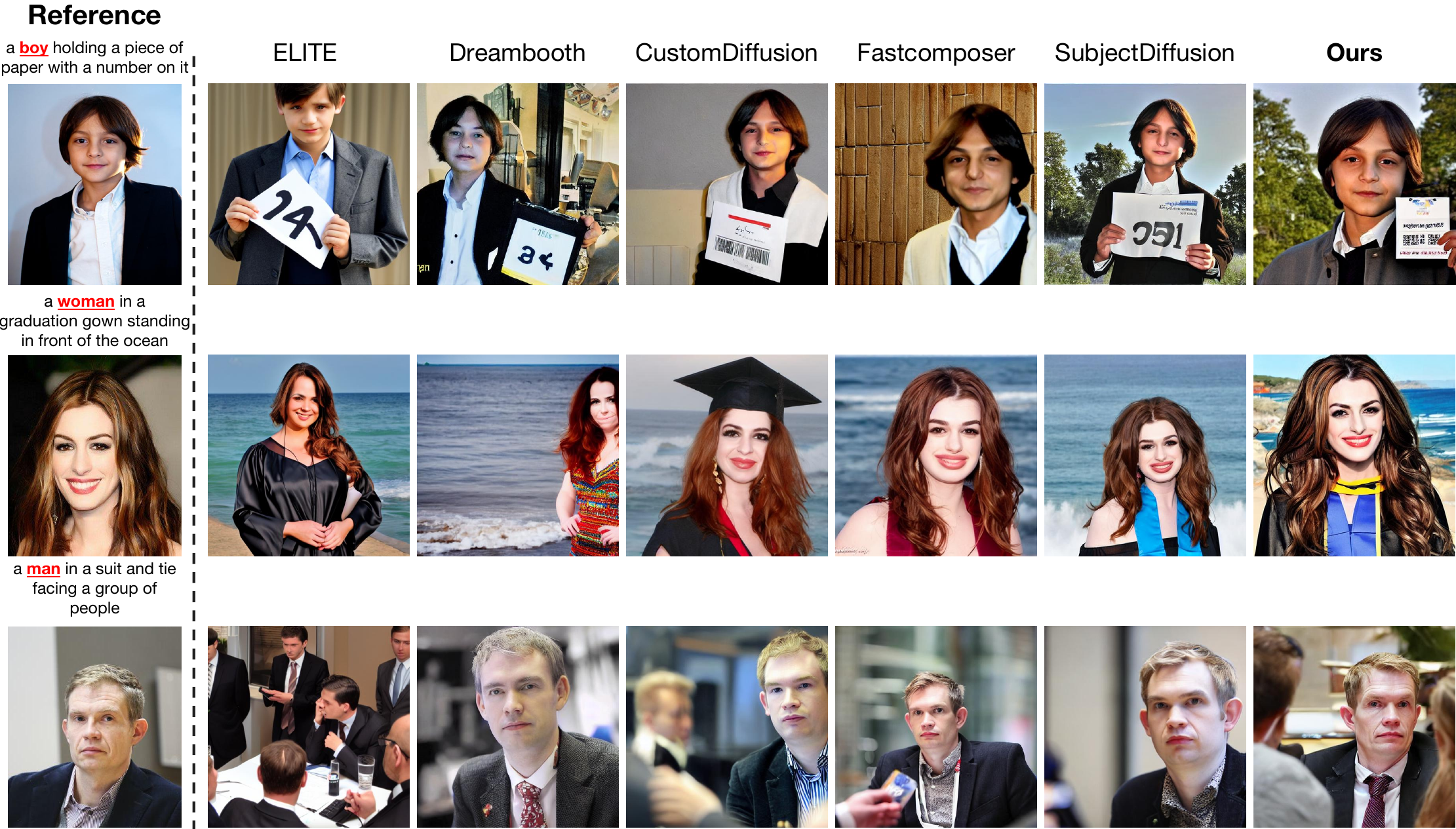}
    \caption{Qualitative comparative results against state-of-the-art methods on single-subject generation.}
    \label{fig:single_compare}
    \vspace{-0.3cm}
\end{figure*}

\begin{figure}[t]
    \centering
    \includegraphics[width=1\linewidth]{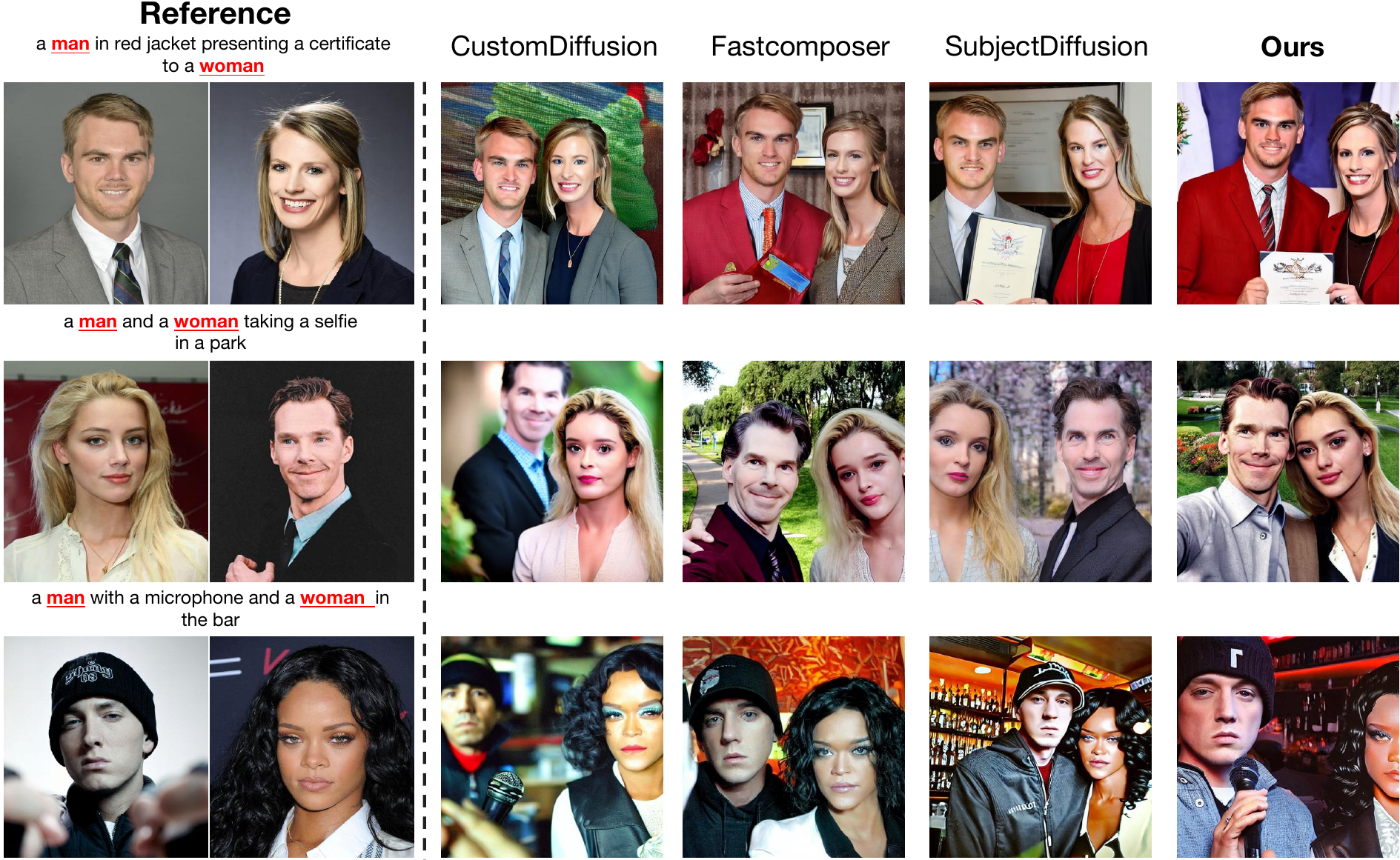}
    \caption{Qualitative comparative results against state-of-the-art methods on multi-subject generation.}
    \label{fig:multi_compare}

\end{figure}
\begin{figure*}[t]
    \centering
    \includegraphics[width=0.8\linewidth]{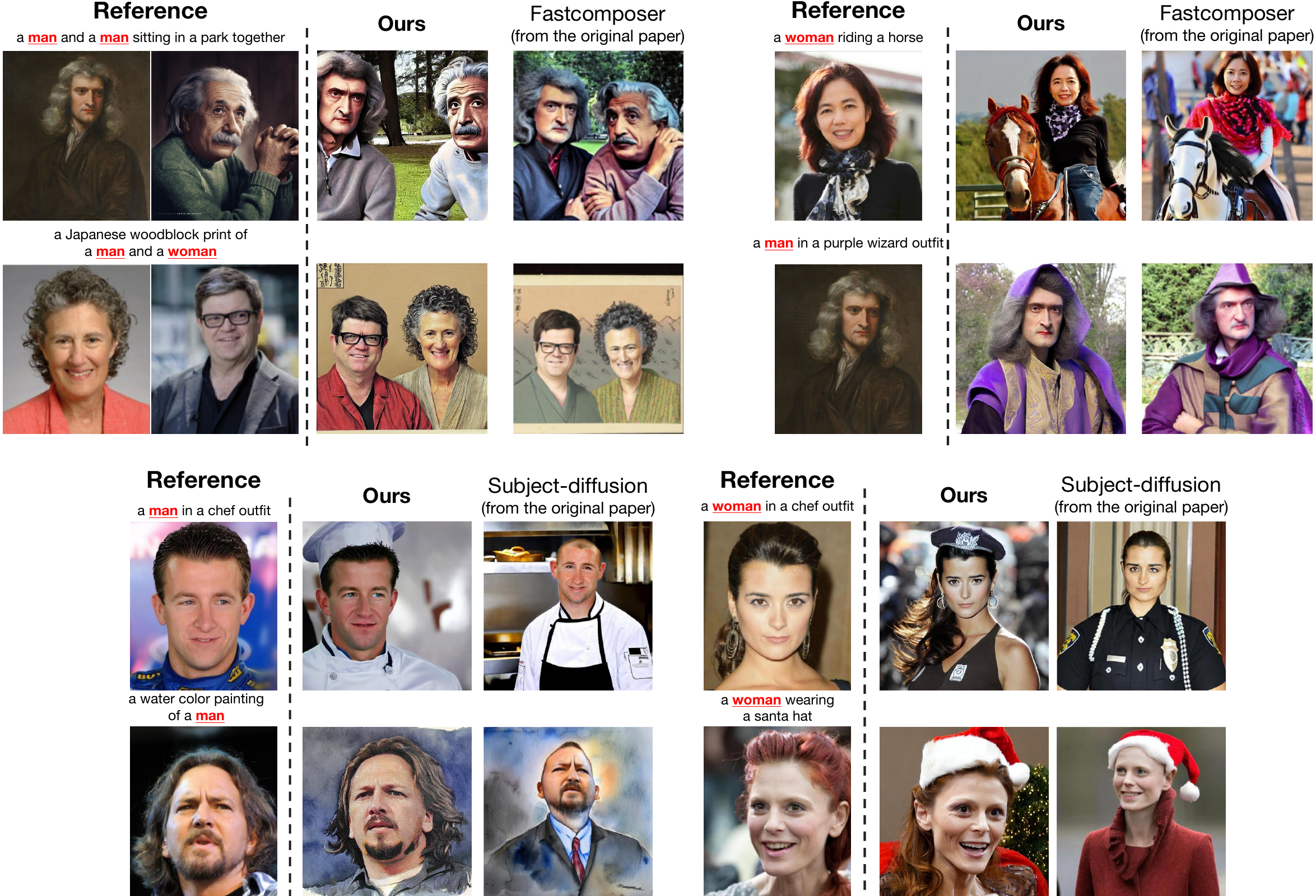}
    \caption{Qualitative comparative results against Fastcomposer and Subject-diffusion using the samples provided in their original papers. The generated results of Fastcomposer and Subject-diffusion are all from their original papers.}
    \label{fig:original_compare}
\end{figure*}

\begin{figure*}[t]
    \centering
    \includegraphics[width=0.8\linewidth]{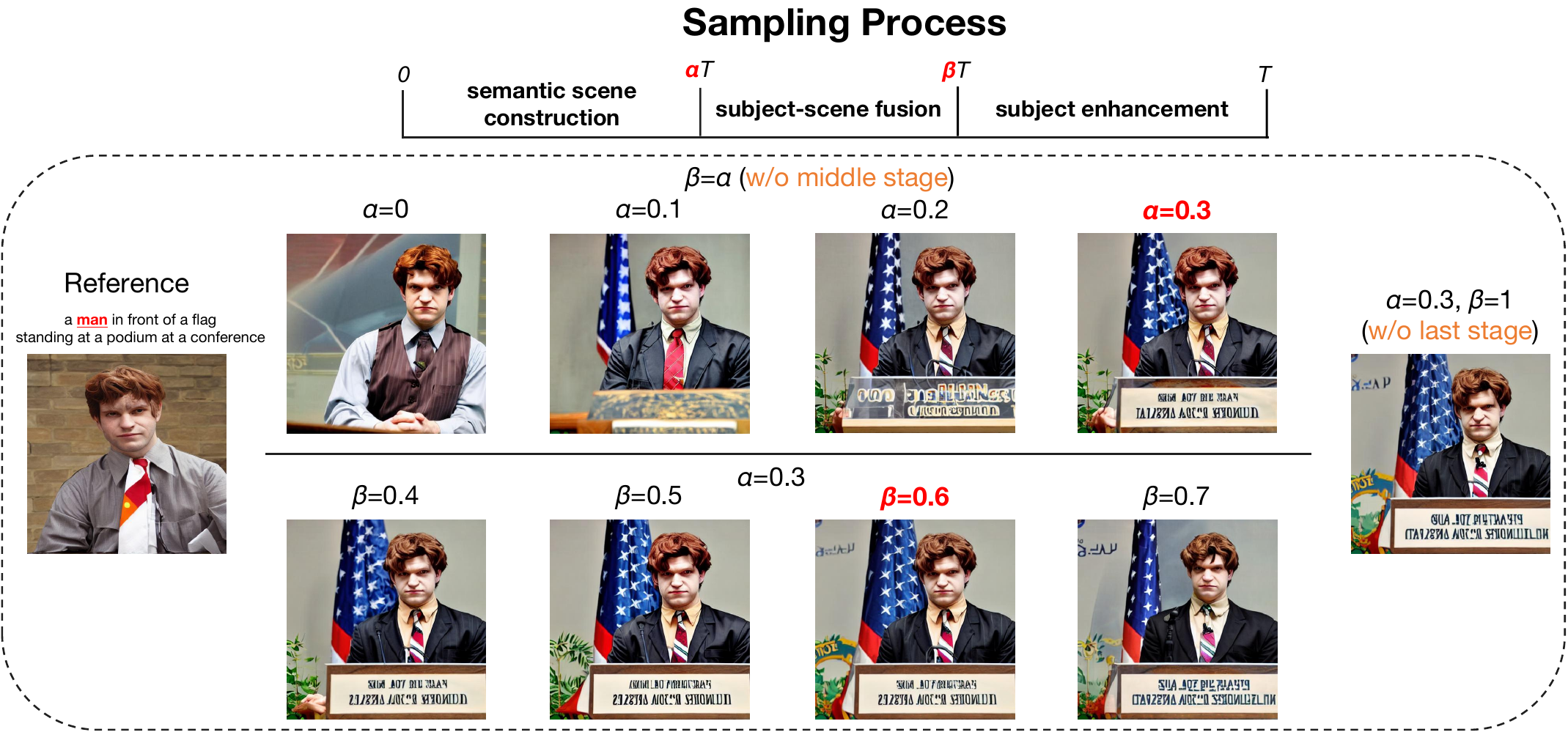}
    \caption{Visualized analysis of hyper-parameters $\alpha$ and $\beta$. We initially set $\beta$ equal to $\alpha$ to explore optimal values for $\alpha$. Once $\alpha$ is determined, we then vary $\beta$ to determine its value. We also intentionally set $\beta$ = 1 to assess the effectiveness of the last stage.}
    \label{fig:effectiveness}
\end{figure*}

\begin{figure}[t]
    \centering
    \includegraphics[width=0.9\linewidth]{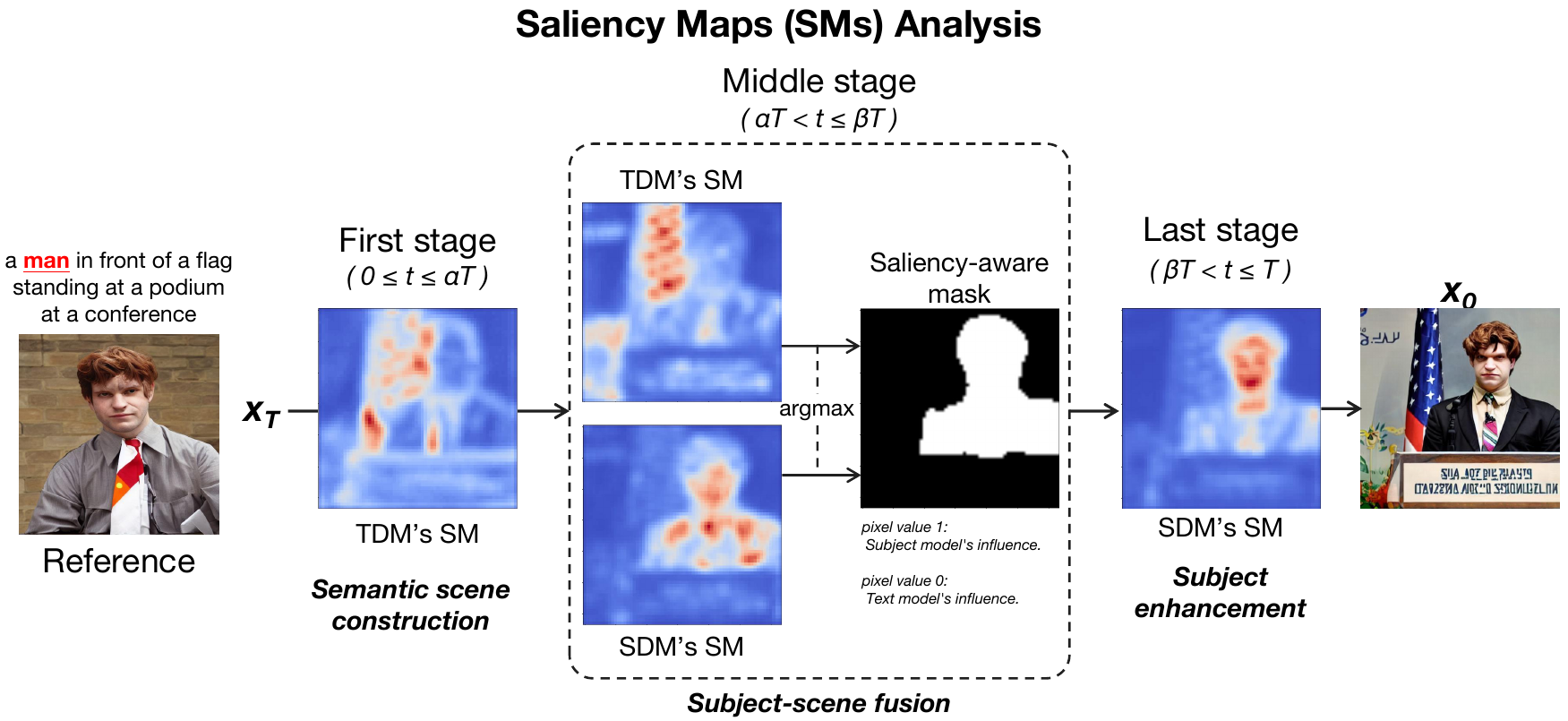}
    \caption{The visualization of average salience maps of our model in each sampling stage.}
    \label{fig:visualize}
    \vspace{-0.4cm}
\end{figure}
\section{Experiment}
\label{sec:exp}
\subsection{Implementation details}
\textbf{\textit{Datasets.}} 
Two datasets are used in our experiments. One is FFHQ-face \cite{fastcomposer}, which is constructed based on the FFHQ-wild dataset \cite{karras2019style} and comprised of a total of 70,000 samples, 60,000 for training, and 10,000 for testing.  Each sample contains a caption and one or more persons. The other dataset is the Single-benchmark dataset employed in the recent work \cite{subject-diffusion, fastcomposer}. It comprises 15 subjects, each with 30 text prompts.

\textbf{\textit{Training Configurations.}} We train our models SDM and TDM based on the pre-trained Stable-Diffusion (SD) v1-5 model \cite{ldm} with the FFHQ-face dataset.  For SDM's image encoder, we utilized OpenAI’s \textit{clip-vit-large-patch14} vision model, which acts as the companion model to the text encoder in SDv1-5. We conducted training on the SDM for a total of 450k steps and on the TDM for 250K steps, utilizing 4 NVIDIA A100 GPUs. We set a constant learning rate of 1e-5 and a batch size of 8. 

\textbf{\textit{Evaluation.}} Quantitively, we evaluate the single-subject generation on Single-benchmark and the multi-subject generation on the test set of FFHQ-face. To further assess the robustness and superiority of our method, we use a diverse set of unseen persons and semantic scenes for qualitative comparison. We assess image generation quality using two metrics: identity preservation (IP) and prompt consistency (PC). 
IP score is obtained by face detection in both reference and generated images using MTCNN \cite{zhang2016joint} and then pairwise identity similarity is calculated using FaceNet \cite{schroff2015facenet}. For multi-subject evaluation, we identified and matched faces in generated images with reference subjects, measuring overall identity preservation based on the minimum similarity value.
PC is assessed through the average CLIP-L/14 image-text similarity, following the approach in textual-inversion \cite{textual_inversion}.

\subsection{Results}
In \cref{table:compare}, we present a quantitative analysis comparing the performance of Face-Diffuser against baseline methods. Our method exhibits a significant advantage in both single-subject and multi-subject image generation. Notably, when compared to the previous state-of-the-art model Fastcomposer on multi-subject generation, Face-diffuser outperforms it by 0.132 and 0.084 in terms of identity preservation and prompt consistency, respectively. While Subject-diffusion and Fastcomposer excel in identity preservation compared to other methods, except our Face-diffuser, they tend to perform less satisfactorily in terms of prompt consistency, potentially due to their overfitting to the text prompts.

Additionally, we provide some qualitative comparisons. As illustrated in \cref{fig:single_compare} and \cref{fig:multi_compare}, our method excels in synthesizing more consistent persons with given reference images and semantic scenes compared to other baselines. For instance, in \cref{fig:single_compare} (1st row), Fastcomposer and CustomDiffusion fail to generate the boy holding a piece of paper. Similarly, in \cref{fig:single_compare} (3rd row), all compared methods except ELITE and CustomDiffusion, struggle to generate the scene of a man facing a group of people, but the two methods also produce unsatisfactory persons.
In \cref{fig:multi_compare}, it is evident that all compared methods fall short of producing satisfactory scenes.

Furthermore, we conduct additional comparisons with the two most recent methods, Fastcomposer and Subject-diffusion, using the samples provided in their original papers for further qualitative evaluation. The visualization results can be observed in \cref{fig:original_compare}, providing additional evidence of our method's superiority in high-fidelity image generation.

\subsection{Hyperparameter and Effectiveness Analysis} \label{effectiveness}
\textbf{Choice of $\alpha$ and $\beta$.} We explore the optimal values for $\alpha$ and $\beta$ in our work since they significantly impact the overall performance of our method. \cref{fig:effectiveness} shows the impact of varying the ratio of $\alpha$ and $\beta$. 

Initially, we prioritize investigating the value of $\alpha$ and set $\beta = \alpha$ (ablating the middle stage) to explore the optimal timesteps for semantic scene construction (\cref{fig:effectiveness} (first row)). We observe that when $\alpha = \beta = 0$, it leads to subject overfitting \cite{fastcomposer}, which is expected due to the absence of TDM. As we increase the ratio, the generated semantic scene improves. When $\alpha = 0.3$, the scene is effectively constructed. However, we encounter issues related to coherence between the character and the scene, highlighting the importance of the middle stage.

Subsequently, we set $\alpha = 0.3$ to determine the value of $\beta$, which represents the sampling steps assigned to the semantic-scene fusion stage (\cref{fig:effectiveness} (second row)). As we increase the ratio, the coherence issue between the person and the scene improves, along with the overall quality of person generation. Optimal fusion between the person and the scene is achieved at $\beta = 0.6$, leading us to determine $\beta = 0.6$.

To further assess the significance of the subject enhancement stage, we set $\alpha = 0.3$ and $\beta = 1$ (removing the last stage). By observing the results, the fidelity of character generation deteriorates. This underscores the substantial role of the subject enhancement stage in enhancing fidelity.

\textbf{Visualization of Salience-adaptive Mask.} We further analyze the effectiveness of our collaborative sampling process by visualizing the average salience map of each model, as shown in \cref{fig:visualize}. These visualizations reveal that TDM mainly focuses on the semantic scene construction, while SDM places its emphasis on subject generation. This alignment with our design philosophy reinforces the distinct roles and responsibilities assigned to each model within the Face-diffuser framework.
\begin{table}[]  \centering
	\small
	\caption{Quantitative results on single- and multi-subject generation. IP denotes identity reservation and PC denotes prompt consistency. "N.A." indicates that the information is not available.}
	\vspace{-0.4cm}
	\label{table1}
	\begin{tabular}{c|cc|ccl}
		\bottomrule
		\multirow{2}{*}{\textbf{Methods}} & \multicolumn{2}{c|}{\textit{\textbf{Single-Subject}}} & \multicolumn{2}{c}{\textit{\textbf{Multi-Subject}}} \\
		& IP  $\uparrow$       & PC $\uparrow$     &IP $\uparrow$      & \multicolumn{2}{c}{PC $\uparrow$}     \\
		\bottomrule
            ELITE (zero-shot)                  & 0.228             & 0.146   &N.A.      & N.A.         \\
		Dreambooth  (fine-tune)                 & 0.273             & 0.239   &N.A.      & N.A.        \\
	     Custom-Diffusion (fine-tune)        & 0.434             & 0.233    &N.A.     & N.A.    \\
		Subject-Diffusion(zero-shot)   & 0.605             & 0.228   &0.432      & 0.205     \\
		Fastcomposer (zero-shot)           & 0.514             & 0.243      &0.461   &0.235     \\

            \bottomrule
		\textbf{Face-Diffuser} (zero-shot)                & \textbf{0.708}     &\textbf{0.325}        & \textbf{0.593}         & \multicolumn{2}{c}{\textbf{0.319}}     \\
		\bottomrule
	\end{tabular} \\

	\label{table:compare}
 \vspace{-0.5cm}
\end{table}

\section{Conclusion}
\label{sec:conclusion}
Current subject-driven person-centric image generation models jointly learn the semantic scene and person generation by fine-tuning a common pre-trained diffusion model, which leads to training imbalance and quality compromises. In this paper, we propose Face-diffuser, an effective collaborative generation pipeline that develops two independent diffusion models for semantic scenes and person generation. The sampling process is divided into three sequential stages, i.e., semantic scene construction, subject-scene fusion, and subject enhancement. The subject-scene fusion stage, that is the collaboration achieved through our novel and highly effective mechanism, Saliency-adaptive Noise Fusion, which spatial blending of predicted noises from both models automatically in a saliency-aware manner in each step. Extensive experiments demonstrate Face-diffuser's effectiveness in generating high-quality images depicting multiple unseen persons in various scenarios.
\section{Acknowledgments}
This work was supported by National Natural Science Fund of China (62176064) and the Key Program of Natural Science Foundation of Zhejiang (LZ24F030012). Computations were performed using Fudan University's CFFF platform. Cheng Jin and Jianwei Zheng are the corresponding authors.
{
    \small
    \bibliographystyle{ieeenat_fullname}
    \bibliography{11_references}

\begin{thebibliography}{37}
\providecommand{\natexlab}[1]{#1}
\providecommand{\url}[1]{\texttt{#1}}
\expandafter\ifx\csname urlstyle\endcsname\relax
  \providecommand{\doi}[1]{doi: #1}\else
  \providecommand{\doi}{doi: \begingroup \urlstyle{rm}\Url}\fi

\bibitem[Avrahami et~al.(2023{\natexlab{a}})Avrahami, Aberman, Fried, Cohen-Or,
  and Lischinski]{avrahami2023break}
Omri Avrahami, Kfir Aberman, Ohad Fried, Daniel Cohen-Or, and Dani Lischinski.
\newblock Break-a-scene: Extracting multiple concepts from a single image.
\newblock \emph{arXiv preprint arXiv:2305.16311}, 2023{\natexlab{a}}.

\bibitem[Avrahami et~al.(2023{\natexlab{b}})Avrahami, Hayes, Gafni, Gupta,
  Taigman, Parikh, Lischinski, Fried, and Yin]{avrahami2023spatext}
Omri Avrahami, Thomas Hayes, Oran Gafni, Sonal Gupta, Yaniv Taigman, Devi
  Parikh, Dani Lischinski, Ohad Fried, and Xi Yin.
\newblock Spatext: Spatio-textual representation for controllable image
  generation.
\newblock In \emph{CVPR}, pages 18370--18380, 2023{\natexlab{b}}.

\bibitem[Bolya and Hoffman(2023)]{bolya2023token}
Daniel Bolya and Judy Hoffman.
\newblock Token merging for fast stable diffusion.
\newblock In \emph{CVPR}, pages 4598--4602, 2023.

\bibitem[Chang et~al.(2023)Chang, Zhang, Barber, Maschinot, Lezama, Jiang,
  Yang, Murphy, Freeman, Rubinstein, et~al.]{chang2023muse}
Huiwen Chang, Han Zhang, Jarred Barber, AJ Maschinot, Jose Lezama, Lu Jiang,
  Ming-Hsuan Yang, Kevin Murphy, William~T Freeman, Michael Rubinstein, et~al.
\newblock Muse: Text-to-image generation via masked generative transformers.
\newblock \emph{arXiv preprint arXiv:2301.00704}, 2023.

\bibitem[Chen et~al.(2022)Chen, Hu, Saharia, and Cohen]{chen2022re}
Wenhu Chen, Hexiang Hu, Chitwan Saharia, and William~W Cohen.
\newblock Re-imagen: Retrieval-augmented text-to-image generator.
\newblock \emph{arXiv preprint arXiv:2209.14491}, 2022.

\bibitem[Chen et~al.(2023)Chen, Hu, Li, Rui, Jia, Chang, and
  Cohen]{chen2023subject}
Wenhu Chen, Hexiang Hu, Yandong Li, Nataniel Rui, Xuhui Jia, Ming-Wei Chang,
  and William~W Cohen.
\newblock Subject-driven text-to-image generation via apprenticeship learning.
\newblock \emph{arXiv preprint arXiv:2304.00186}, 2023.

\bibitem[Ding et~al.(2021)Ding, Yang, Hong, Zheng, Zhou, Yin, Lin, Zou, Shao,
  Yang, et~al.]{ding2021cogview}
Ming Ding, Zhuoyi Yang, Wenyi Hong, Wendi Zheng, Chang Zhou, Da Yin, Junyang
  Lin, Xu Zou, Zhou Shao, Hongxia Yang, et~al.
\newblock Cogview: Mastering text-to-image generation via transformers.
\newblock In \emph{NeruIPS}, pages 19822--19835, 2021.

\bibitem[Feng et~al.(2023{\natexlab{a}})Feng, Jiang, Xu, and Zheng]{DMINet}
Yuchao Feng, Jiawei Jiang, Honghui Xu, and Jianwei Zheng.
\newblock Change detection on remote sensing images using dual-branch
  multilevel intertemporal network.
\newblock \emph{IEEE Transactions on Geoscience and Remote Sensing},
  61:\penalty0 1--15, 2023{\natexlab{a}}.

\bibitem[Feng et~al.(2023{\natexlab{b}})Feng, Shao, Xu, Xu, and Zheng]{LCANet}
Yuchao Feng, Yanyan Shao, Honghui Xu, Jinshan Xu, and Jianwei Zheng.
\newblock A lightweight collective-attention network for change detection.
\newblock In \emph{ACM MM}, page 8195–8203, 2023{\natexlab{b}}.

\bibitem[Gal et~al.(2023{\natexlab{a}})Gal, Alaluf, Atzmon, Patashnik, Bermano,
  Chechik, and Cohen-Or]{textual_inversion}
Rinon Gal, Yuval Alaluf, Yuval Atzmon, Or Patashnik, Amit~H Bermano, Gal
  Chechik, and Daniel Cohen-Or.
\newblock An image is worth one word: Personalizing text-to-image generation
  using textual inversion.
\newblock In \emph{ICLR}, 2023{\natexlab{a}}.

\bibitem[Gal et~al.(2023{\natexlab{b}})Gal, Arar, Atzmon, Bermano, Chechik, and
  Cohen-Or]{gal2023designing}
Rinon Gal, Moab Arar, Yuval Atzmon, Amit~H Bermano, Gal Chechik, and Daniel
  Cohen-Or.
\newblock Designing an encoder for fast personalization of text-to-image
  models.
\newblock In \emph{Siggraph}, 2023{\natexlab{b}}.

\bibitem[Hao et~al.(2023)Hao, Han, Zhao, and Wong]{hao2023vico}
Shaozhe Hao, Kai Han, Shihao Zhao, and Kwan-Yee~K Wong.
\newblock Vico: Detail-preserving visual condition for personalized
  text-to-image generation.
\newblock \emph{arXiv preprint arXiv:2306.00971}, 2023.

\bibitem[Kang et~al.(2023)Kang, Zhu, Zhang, Park, Shechtman, Paris, and
  Park]{kang2023scaling}
Minguk Kang, Jun-Yan Zhu, Richard Zhang, Jaesik Park, Eli Shechtman, Sylvain
  Paris, and Taesung Park.
\newblock Scaling up gans for text-to-image synthesis.
\newblock In \emph{CVPR}, pages 10124--10134, 2023.

\bibitem[Karras et~al.(2019)Karras, Laine, and Aila]{karras2019style}
Tero Karras, Samuli Laine, and Timo Aila.
\newblock A style-based generator architecture for generative adversarial
  networks.
\newblock In \emph{CVPR}, pages 4401--4410, 2019.

\bibitem[Kirstain et~al.(2023)Kirstain, Levy, and Polyak]{kirstain2023x}
Yuval Kirstain, Omer Levy, and Adam Polyak.
\newblock X\&fuse: Fusing visual information in text-to-image generation.
\newblock \emph{arXiv preprint arXiv:2303.01000}, 2023.

\bibitem[Kumari et~al.(2023)Kumari, Zhang, Zhang, Shechtman, and
  Zhu]{custom-diffusion}
Nupur Kumari, Bingliang Zhang, Richard Zhang, Eli Shechtman, and Jun-Yan Zhu.
\newblock Multi-concept customization of text-to-image diffusion.
\newblock In \emph{CVPR}, pages 1931--1941, 2023.

\bibitem[Ma et~al.(2023)Ma, Liang, Chen, and Lu]{subject-diffusion}
Jian Ma, Junhao Liang, Chen Chen, and Haonan Lu.
\newblock Subject-diffusion: Open domain personalized text-to-image generation
  without test-time fine-tuning.
\newblock \emph{arXiv preprint arXiv:2307.11410}, 2023.

\bibitem[Roich et~al.(2022)Roich, Mokady, Bermano, and
  Cohen-Or]{tuning-encoder}
Daniel Roich, Ron Mokady, Amit~H Bermano, and Daniel Cohen-Or.
\newblock Pivotal tuning for latent-based editing of real images.
\newblock \emph{ACM Transactions on Graphics (TOG)}, 42\penalty0 (1):\penalty0
  1--13, 2022.

\bibitem[Rombach et~al.(2022)Rombach, Blattmann, Lorenz, Esser, and Ommer]{ldm}
Robin Rombach, Andreas Blattmann, Dominik Lorenz, Patrick Esser, and Bj{\"o}rn
  Ommer.
\newblock High-resolution image synthesis with latent diffusion models.
\newblock In \emph{CVPR}, pages 10684--10695, 2022.

\bibitem[Ruiz et~al.(2023)Ruiz, Li, Jampani, Pritch, Rubinstein, and
  Aberman]{dreambooth}
Nataniel Ruiz, Yuanzhen Li, Varun Jampani, Yael Pritch, Michael Rubinstein, and
  Kfir Aberman.
\newblock Dreambooth: Fine tuning text-to-image diffusion models for
  subject-driven generation.
\newblock In \emph{CVPR}, pages 22500--22510, 2023.

\bibitem[Saharia et~al.(2022)Saharia, Chan, Saxena, Li, Whang, Denton,
  Ghasemipour, Gontijo~Lopes, Karagol~Ayan, Salimans,
  et~al.]{saharia2022photorealistic}
Chitwan Saharia, William Chan, Saurabh Saxena, Lala Li, Jay Whang, Emily~L
  Denton, Kamyar Ghasemipour, Raphael Gontijo~Lopes, Burcu Karagol~Ayan, Tim
  Salimans, et~al.
\newblock Photorealistic text-to-image diffusion models with deep language
  understanding.
\newblock In \emph{NeurIPS}, pages 36479--36494, 2022.

\bibitem[Sarukkai et~al.(2023)Sarukkai, Li, Ma, R{\'e}, and
  Fatahalian]{sarukkai2023collage}
Vishnu Sarukkai, Linden Li, Arden Ma, Christopher R{\'e}, and Kayvon
  Fatahalian.
\newblock Collage diffusion.
\newblock \emph{arXiv preprint arXiv:2303.00262}, 2023.

\bibitem[Schroff et~al.(2015)Schroff, Kalenichenko, and
  Philbin]{schroff2015facenet}
Florian Schroff, Dmitry Kalenichenko, and James Philbin.
\newblock Facenet: A unified embedding for face recognition and clustering.
\newblock In \emph{CVPR}, pages 815--823, 2015.

\bibitem[Schuhmann et~al.(2022)Schuhmann, Beaumont, Vencu, Gordon, Wightman,
  Cherti, Coombes, Katta, Mullis, Wortsman, et~al.]{schuhmann2022laion}
Christoph Schuhmann, Romain Beaumont, Richard Vencu, Cade Gordon, Ross
  Wightman, Mehdi Cherti, Theo Coombes, Aarush Katta, Clayton Mullis, Mitchell
  Wortsman, et~al.
\newblock Laion-5b: An open large-scale dataset for training next generation
  image-text models.
\newblock In \emph{NeurIPS}, pages 25278--25294, 2022.

\bibitem[Shi et~al.(2023)Shi, Xiong, Lin, and Jung]{shi2023instantbooth}
Jing Shi, Wei Xiong, Zhe Lin, and Hyun~Joon Jung.
\newblock Instantbooth: Personalized text-to-image generation without test-time
  finetuning.
\newblock \emph{arXiv preprint arXiv:2304.03411}, 2023.

\bibitem[Si et~al.(2023)Si, Huang, Jiang, and Liu]{si2023freeu}
Chenyang Si, Ziqi Huang, Yuming Jiang, and Ziwei Liu.
\newblock Freeu: Free lunch in diffusion u-net.
\newblock \emph{arXiv preprint arXiv:2309.11497}, 2023.

\bibitem[Sohl-Dickstein et~al.(2015)Sohl-Dickstein, Weiss, Maheswaranathan, and
  Ganguli]{sohl2015deep}
Jascha Sohl-Dickstein, Eric Weiss, Niru Maheswaranathan, and Surya Ganguli.
\newblock Deep unsupervised learning using nonequilibrium thermodynamics.
\newblock In \emph{ICML}, pages 2256--2265, 2015.

\bibitem[Tewel et~al.(2023)Tewel, Gal, Chechik, and Atzmon]{tewel2023key}
Yoad Tewel, Rinon Gal, Gal Chechik, and Yuval Atzmon.
\newblock Key-locked rank one editing for text-to-image personalization.
\newblock In \emph{ACM SIGGRAPH}, pages 1--11, 2023.

\bibitem[Vaswani et~al.(2017)Vaswani, Shazeer, Parmar, Uszkoreit, Jones, Gomez,
  Kaiser, and Polosukhin]{vaswani2017attention}
Ashish Vaswani, Noam Shazeer, Niki Parmar, Jakob Uszkoreit, Llion Jones,
  Aidan~N Gomez, {\L}ukasz Kaiser, and Illia Polosukhin.
\newblock Attention is all you need.
\newblock In \emph{NeurIPS}, 2017.

\bibitem[von Platen et~al.(2022)von Platen, Patil, Lozhkov, Cuenca, Lambert,
  Rasul, Davaadorj, and Wolf]{von-platen-etal-2022-diffusers}
Patrick von Platen, Suraj Patil, Anton Lozhkov, Pedro Cuenca, Nathan Lambert,
  Kashif Rasul, Mishig Davaadorj, and Thomas Wolf.
\newblock Diffusers: State-of-the-art diffusion models.
\newblock \url{https://github.com/huggingface/diffusers}, 2022.

\bibitem[Wang et~al.(2023{\natexlab{a}})Wang, Feng, Wu, Xu, and
  Zheng]{wang2023gan}
Yibin Wang, Yuchao Feng, Jie Wu, Honghui Xu, and Jianwei Zheng.
\newblock Ca-gan: Object placement via coalescing attention based generative
  adversarial network.
\newblock In \emph{ICME}, pages 2375--2380, 2023{\natexlab{a}}.

\bibitem[Wang et~al.(2023{\natexlab{b}})Wang, Long, Zhou, Bo, and
  Zheng]{wang2023plsnet}
Yibin Wang, Haixia Long, Qianwei Zhou, Tao Bo, and Jianwei Zheng.
\newblock Plsnet: Position-aware gcn-based autism spectrum disorder diagnosis
  via fc learning and rois sifting.
\newblock \emph{Computers in Biology and Medicine}, page 107184,
  2023{\natexlab{b}}.

\bibitem[Wei et~al.(2023)Wei, Zhang, Ji, Bai, Zhang, and Zuo]{wei2023elite}
Yuxiang Wei, Yabo Zhang, Zhilong Ji, Jinfeng Bai, Lei Zhang, and Wangmeng Zuo.
\newblock Elite: Encoding visual concepts into textual embeddings for
  customized text-to-image generation.
\newblock \emph{arXiv preprint arXiv:2302.13848}, 2023.

\bibitem[Xiao et~al.(2023)Xiao, Yin, Freeman, Durand, and Han]{fastcomposer}
Guangxuan Xiao, Tianwei Yin, William~T Freeman, Fr{\'e}do Durand, and Song Han.
\newblock Fastcomposer: Tuning-free multi-subject image generation with
  localized attention.
\newblock \emph{arXiv preprint arXiv:2305.10431}, 2023.

\bibitem[Zhang et~al.(2016)Zhang, Zhang, Li, and Qiao]{zhang2016joint}
Kaipeng Zhang, Zhanpeng Zhang, Zhifeng Li, and Yu Qiao.
\newblock Joint face detection and alignment using multitask cascaded
  convolutional networks.
\newblock \emph{IEEE signal processing letters}, 23\penalty0 (10):\penalty0
  1499--1503, 2016.

\bibitem[Zhao et~al.(2023)Zhao, Zheng, Wang, Lan, and
  Yang]{zhao2023magicfusion}
Jing Zhao, Heliang Zheng, Chaoyue Wang, Long Lan, and Wenjing Yang.
\newblock Magicfusion: Boosting text-to-image generation performance by fusing
  diffusion models.
\newblock \emph{arXiv preprint arXiv:2303.13126}, 2023.

\bibitem[Zheng et~al.(2023)Zheng, Zhou, Li, Qi, Shan, and
  Li]{zheng2023layoutdiffusion}
Guangcong Zheng, Xianpan Zhou, Xuewei Li, Zhongang Qi, Ying Shan, and Xi Li.
\newblock Layoutdiffusion: Controllable diffusion model for layout-to-image
  generation.
\newblock In \emph{CVPR}, pages 22490--22499, 2023.

\end{thebibliography}
}

 \clearpage \appendix
\label{sec:appendix}
\section{More cases of problems}
\begin{figure*}[t]
    \centering
    \includegraphics[width=\linewidth]{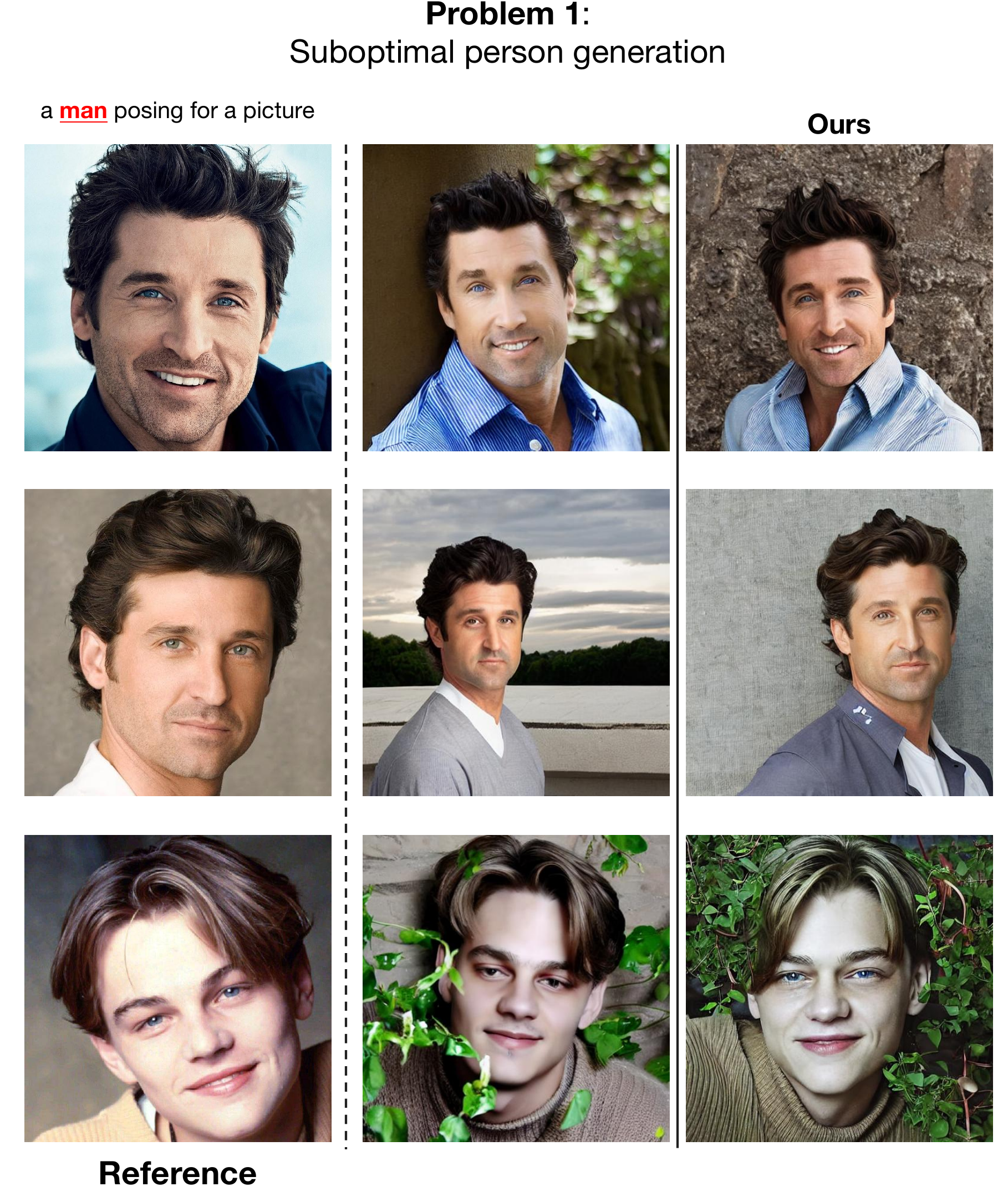}
    \caption{More problem cases of suboptimal person generation.}
    \label{fig:problem1_supply}
\end{figure*}

\begin{figure*}[t]
    \centering
    \includegraphics[width=\linewidth]{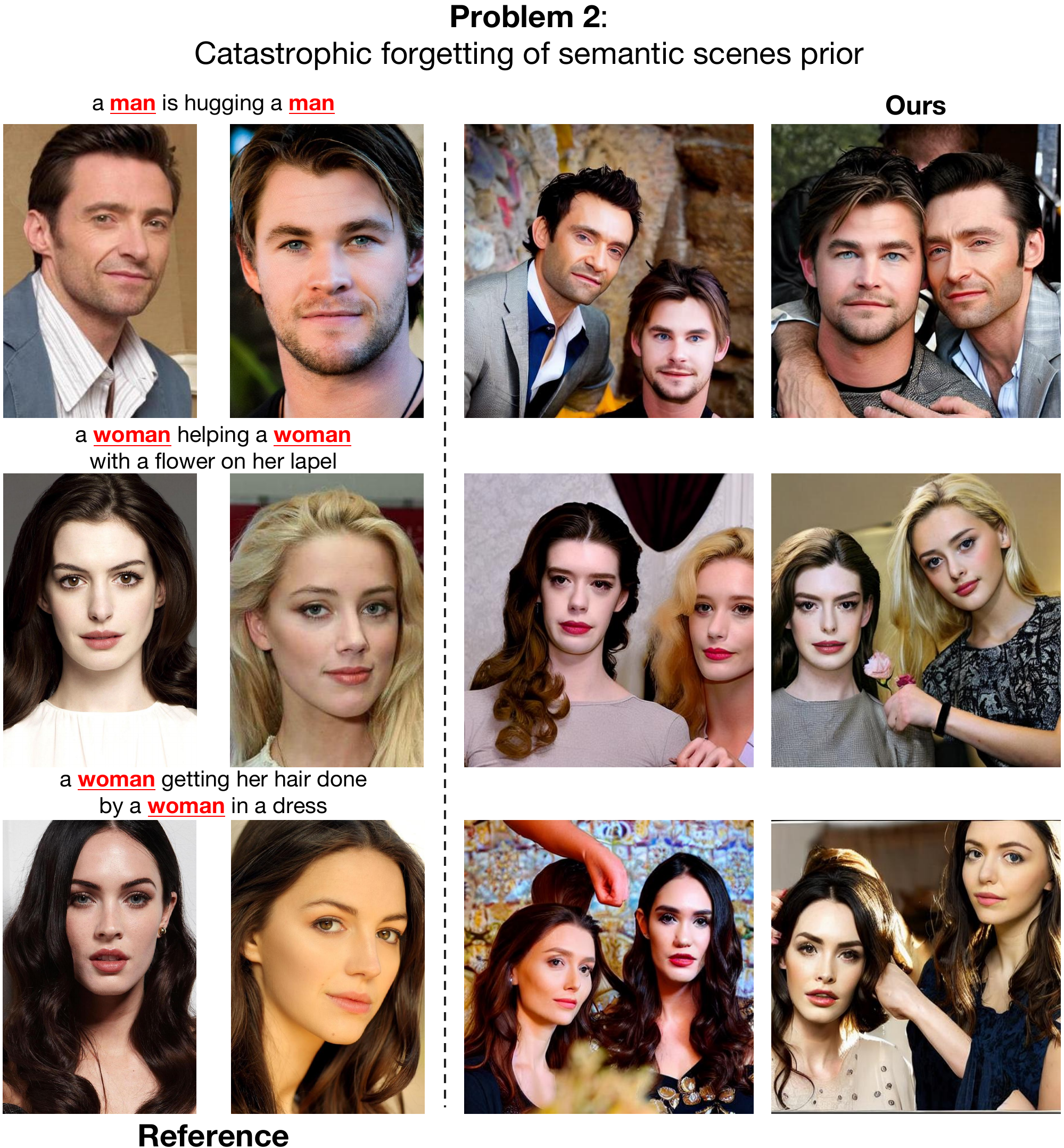}
    \caption{More problem cases of catastrophic forgetting of semantic scenes prior}
    \label{fig:problem2_supply}
\end{figure*}

More cases of the challenges confronted by current SOTA methods are supplied in \cref{fig:problem1_supply} and \cref{fig:problem2_supply}.

\section{Algorithm}
The computation pipeline of Saliency-adaptive Noise Fusion is illustrated in \cref{alg:SNF}.

\begin{algorithm}[!h]
    \caption{SNF}
    \label{alg:SNF}
    \renewcommand{\algorithmicrequire}{\textbf{Input:}}
    \renewcommand{\algorithmicensure}{\textbf{Output:}}
    \begin{algorithmic}[1]
        \REQUIRE TDM $\varepsilon_{\theta_{T}}$, SDM $\varepsilon_{\theta_{S}}$, text prompt $\psi(c)$ and augmented text prompt $\psi(c)_{aug}$, the noise $\textbf{\textit{x}}_{T(1-\alpha)}$. 
        \ENSURE The noise $\textbf{\textit{x}}_{T(1-\beta)}$    
        
        \FOR{each $t$ from $T(1-\alpha)$ to $T(1-\beta)$} 
            \STATE $\bm{\varepsilon}_{T} = \varepsilon_{\theta_{T}}(\textbf{\textit{x}}_{t}|\psi(c))$
            \STATE $\bm{\varepsilon}_{S} = \varepsilon_{\theta_{S}}(\textbf{\textit{x}}_{t}|\psi(c)_{aug})$
            \STATE get $\bm{\mathit{\Omega}}^{T}$ and $\bm{\mathit{\Omega}}^{S}$ via Eq. (3) and Eq. (4)
            \STATE \textbf{\textit{M}} = argmax(Softmax($\bm{\mathit{\Omega}}^{T}$), Softmax($\bm{\mathit{\Omega}}^{S}$))
            \STATE get predicted noises $\hat{\bm{\varepsilon}_{S}}$ and $\hat{\bm{\varepsilon}_{T}}$ via Eq. (2) and Eq. (1)
            \STATE $\hat{\bm{\varepsilon}}$ = \textbf{\textit{M}} $\odot$ $\hat{\bm{\varepsilon}}_{S}$ + (1 - \textbf{\textit{M}}) $\odot$ $\hat{\bm{\varepsilon}}_{T}$
            \STATE $\textbf{\textit{x}}_{t-1} \gets$  $\hat{\bm{\varepsilon}}$ 
        \ENDFOR
        \RETURN $\textbf{\textit{x}}_{T(1-\beta)}$
    \end{algorithmic}
\end{algorithm}

\section{More implementation details}
\textbf{\textit{Baselines.}} We compare with recent state-of-the-art subject-to-image synthesis methods, which included optimization-based techniques like DreamBooth \cite{dreambooth} and Custom-diffusion \cite{custom-diffusion}. These models necessitate subject-specific fine-tuning for each subject. We utilize five images per subject for their fine-tuning in our work. We employed implementations from the diffuser library \cite{von-platen-etal-2022-diffusers} for these methods. Additionally, we also compare with some tuning-free approaches, such as ELITE \cite{wei2023elite}, Subject-diffusion \cite{subject-diffusion}, and Fastcomposer \cite{fastcomposer}. We utilized pre-trained models from the original authors for ELITE and Fastcomposer. However, since Subject-diffusion does not provide a pre-trained model or dataset to the public, we train it on the FFHQ-face \cite{fastcomposer} dataset, adhering to the original paper's settings as closely as possible. Subsequently, we selected its best model for our comparative analysis.

\textbf{\textit{Training Configurations.}} During the training phase, we adopted a strategy following \cite{fastcomposer}, where we freeze the text encoder and only train the U-Net, the MLP module, and the last two transformer blocks of the image encoder. For SDM, we trained only with text condition for 20\% of the samples, a measure taken to preserve the model's capacity for text-only generation. Furthermore, we applied loss functions exclusively within the subject region for half of the training samples, a step taken to enhance the quality of generation in the subject area. Meanwhile, for TDM, we opted for training without any conditions in place for 20\% of the instances, a choice made to facilitate classifier-free guidance sampling. 

\section{More qualitative comparison}
\begin{figure*}[t]
    \centering
    \includegraphics[width=0.8\linewidth]{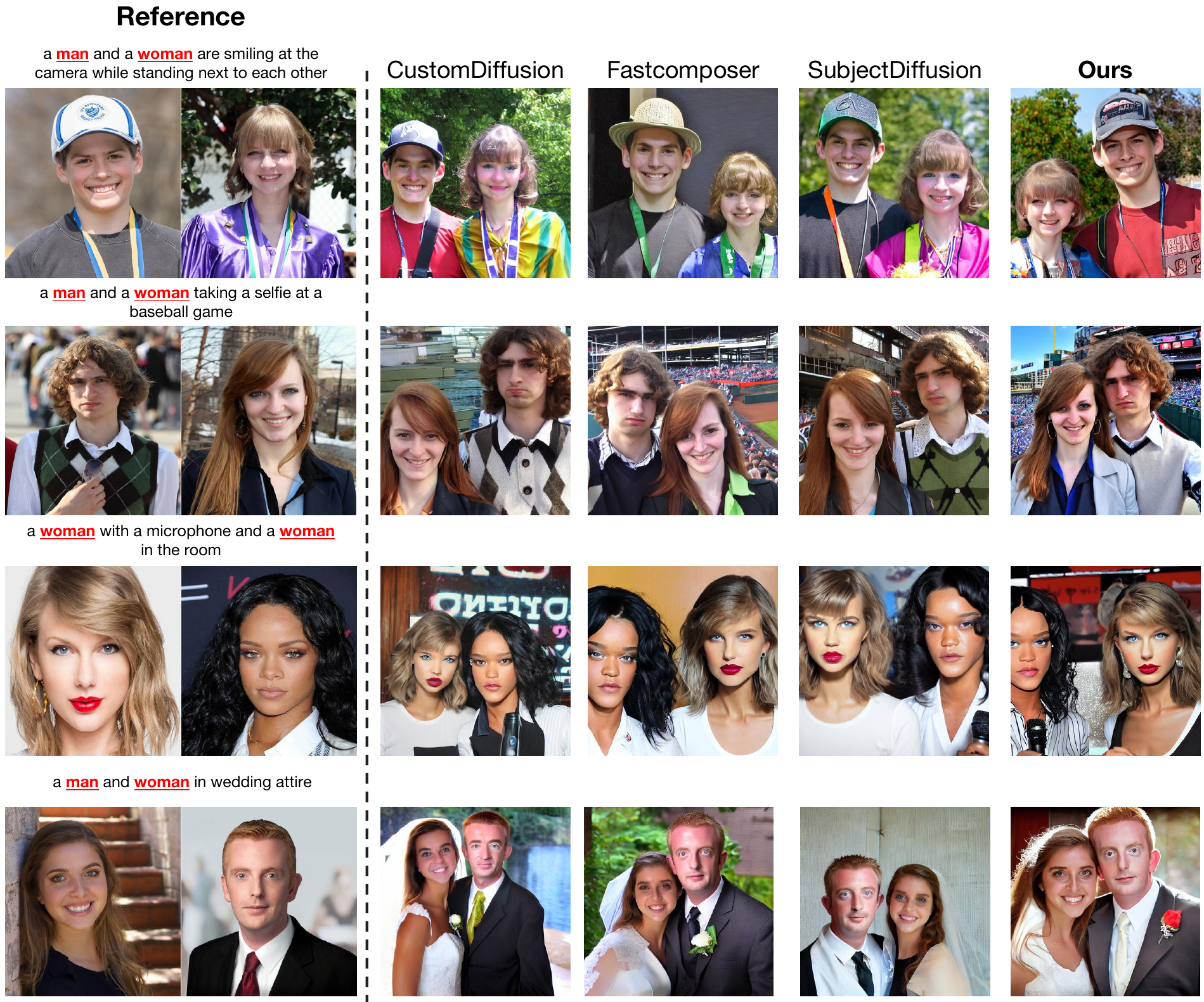}
    \caption{More qualitative comparative results against state-of-the-art methods on multi-subject generation.}
    \label{fig:multi_compare_supply}
\end{figure*}

\begin{figure*}[t]
    \centering
    \includegraphics[width=1\linewidth]{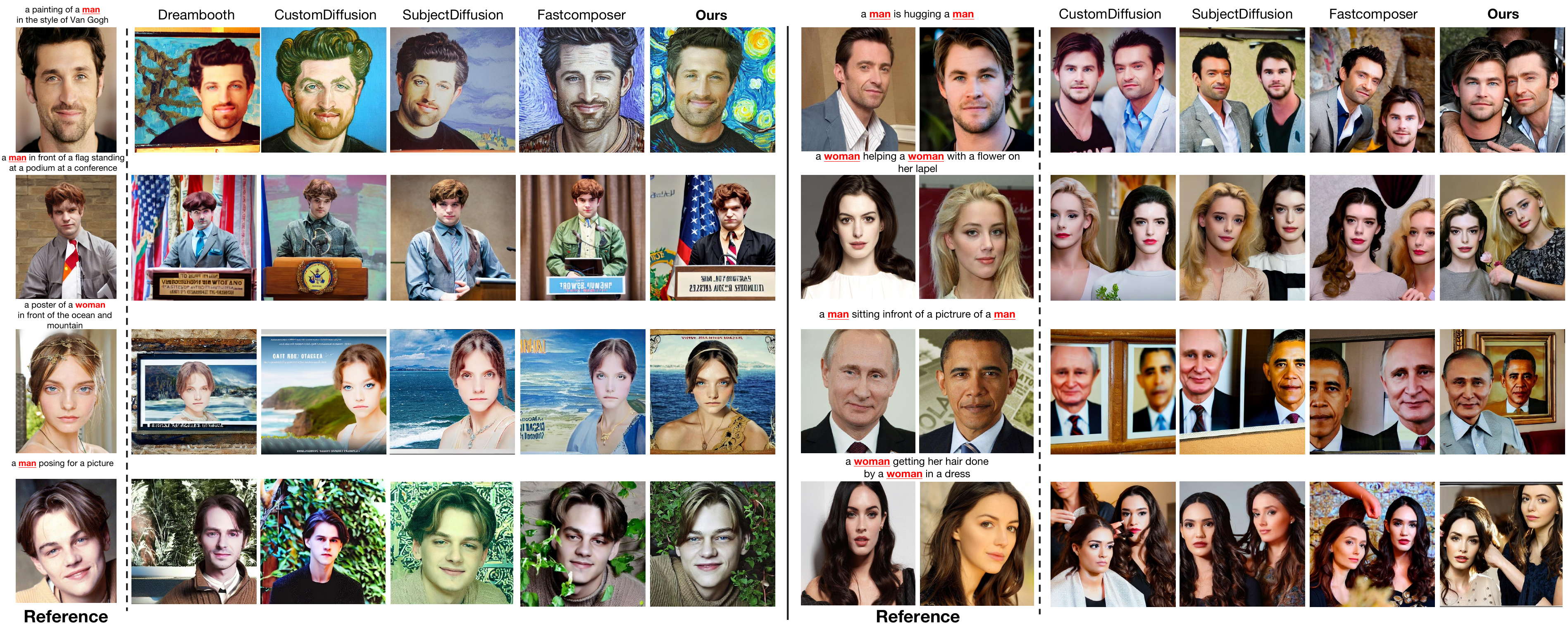}
    \caption{More qualitative comparative results.}
    \label{fig:rebuttal}
\end{figure*}

Additional qualitative comparison results are presented in \cref{fig:multi_compare_supply} and \cref{fig:rebuttal}.

\begin{table}[]  \centering
	 \tiny
	\caption{Additional quantitative comparison results. "N.A." indicates that the information is not available.}
        \tabcolsep=0.1cm
 \vspace{-0.197cm}	
 \label{table1}
	\begin{tabular}{c|cccc|ccccl}
		\bottomrule
		\multirow{2}{*}{\textbf{Methods}} & \multicolumn{4}{c|}{\textit{\textbf{Single-Subject}}} & \multicolumn{4}{c}{\textit{\textbf{Multi-Subject}}} \\
		       & FID $\downarrow$ & IS $\uparrow$  & CLIP-I $\uparrow$     &DINO $\uparrow$    & \multicolumn{1}{c}{FID $\downarrow$} & IS $\uparrow$  & CLIP-I $\uparrow$     &DINO $\uparrow$   \\
		\bottomrule
            ELITE                                & 51.3  &7.83 & 0.722 &  0.571   & N.A.  & N.A. & N.A. &  N.A. \\

            Dreambooth                                & 41.6   & 7.98 & 0.763 & 0.648    & N.A. &N.A.    &N.A. & N.A.  \\

            Custom-Diffusion                                & 35.7  &8.44 & 0.785 & 0.662    & N.A.   & N.A.    & N.A. & N.A.  \\
		
		Subject-Diffusion               & 31.4 &8.92 & 0.778 &   0.727    & 36.7 &7.44 & 0.718 &  0.583   \\
		Fastcomposer                    & 29.8 &9.16 & 0.795 &   0.719   &32.1 & 8.17 & 0.721 & 0.602   \\

            \bottomrule
		\textbf{Face-Diffuser}               & \textbf{21.2}   &\textbf{11.42} & \textbf{0.832} &     \textbf{0.753}         & \multicolumn{1}{c}{\textbf{25.9}}  & \textbf{10.33} &\textbf{0.754} &  \textbf{0.633}  \\
		\bottomrule
	\end{tabular} \\
	\label{table:fid_compare}

 \vspace{-0.2cm}
\end{table}

\section{More quantitative comparison}
Additional quantitative comparison results are presented in \cref{table:fid_compare}.

\section{Ablation study}
\textbf{\textit{The functionality of three sampling stages.}}
We conducte ablation experiments to assess the effectiveness of each stage by removing them individually. The results, as presented in \cref{table:ablation_stage}, highlight the significance of each stage. Removing the semantic scene construction stage notably affects prompt consistency, indicating its role in generating an initial layout for subsequent stages, thus ensuring overall semantic consistency in the generated images. The absence of the subject-scene fusion stage leads to a substantial drop in prompt consistency, emphasizing its importance in maintaining coherence between subjects and scenes, ultimately impacting image fidelity. Additionally, removing the subject enhancement stage resulted in a significant decrease in identity preservation performance, underscoring its role in enhancing the fidelity of generated persons.

\begin{table}[]  \centering
	\scriptsize
	\caption{The quantitative results for ablating each stage on both single- and multi-subject generation tasks. IP denotes identity reservation and PC denotes prompt consistency.}
	\vspace{-0.4cm}
	\label{table1}
	\begin{tabular}{c|cc|ccl}
		\bottomrule
		\multirow{2}{*}{\textbf{Methods}} & \multicolumn{2}{c|}{\textit{\textbf{Single-Subject}}} & \multicolumn{2}{c}{\textit{\textbf{Multi-Subject}}} \\
		& IP  $\uparrow$       & PC $\uparrow$     &IP $\uparrow$      & \multicolumn{2}{c}{PC $\uparrow$}     \\
		\bottomrule
	     w/o semantic scene construction        & 0.699             & 0.268    &0.587     & 0.235     \\
		w/o subject-scene fusion   & \textbf{0.710}             & 0.244   &0.588      & 0.229     \\
		w/o subject enhancement           & 0.583             & 0.322      &0.471   &\textbf{0.322}     \\
		\textbf{Face-Diffuser}                & 0.708     &\textbf{0.325}        & \textbf{0.593}         & \multicolumn{2}{c}{0.319}     \\
		\bottomrule
	\end{tabular} \\
	\vspace{-0.4cm}
	\label{table:ablation_stage}
\end{table}

\begin{table}[]  \centering
	\scriptsize
	\caption{The quantitative results for replacing SNF with direct addition of predicted noises from SDM and TDM on both single- and multi-subject generation tasks. IP denotes identity reservation and PC denotes prompt consistency.}
	\vspace{-0.4cm}
	\label{table1}
	\begin{tabular}{c|cc|ccl}
		\bottomrule
		\multirow{2}{*}{\textbf{Methods}} & \multicolumn{2}{c|}{\textit{\textbf{Single-Subject}}} & \multicolumn{2}{c}{\textit{\textbf{Multi-Subject}}} \\
		& IP  $\uparrow$       & PC $\uparrow$     &IP $\uparrow$      & \multicolumn{2}{c}{PC $\uparrow$}     \\
		\bottomrule
		addition           & 0.523             & 0.221      &0.486   &0.207     \\
		\textbf{saliency-adaptive noise fusion}                & \textbf{0.708}     &\textbf{0.325}        & \textbf{0.593}         & \multicolumn{2}{c}{\textbf{0.319}}     \\
		\bottomrule
	\end{tabular} \\
	\vspace{-0.4cm}
	\label{table:ablation_SNF}
\end{table}

\textbf{\textit{The functionality of Saliency-adaptive Noise Fusion.}}
To further underscore the effectiveness of our proposed Saliency-adaptive Noise Fusion (SNF), we conduct ablation experiments by replacing SNF with the direct addition of two predicted noises from SDM and TDM. The results, as presented in Table \cref{table:ablation_SNF}, clearly highlight the pivotal role of SNF in preserving the unique strengths of each model and achieving an effective collaboration between two generators. It is evident that direct addition leads to a significant degradation in both identity preservation and prompt consistency. This outcome is unsurprising, as direct addition disregards the specialized expertise of each model.

\section{More cases of hyper-parameter analysis}
\begin{figure*}[]
    \centering
    \includegraphics[width=0.8\linewidth]{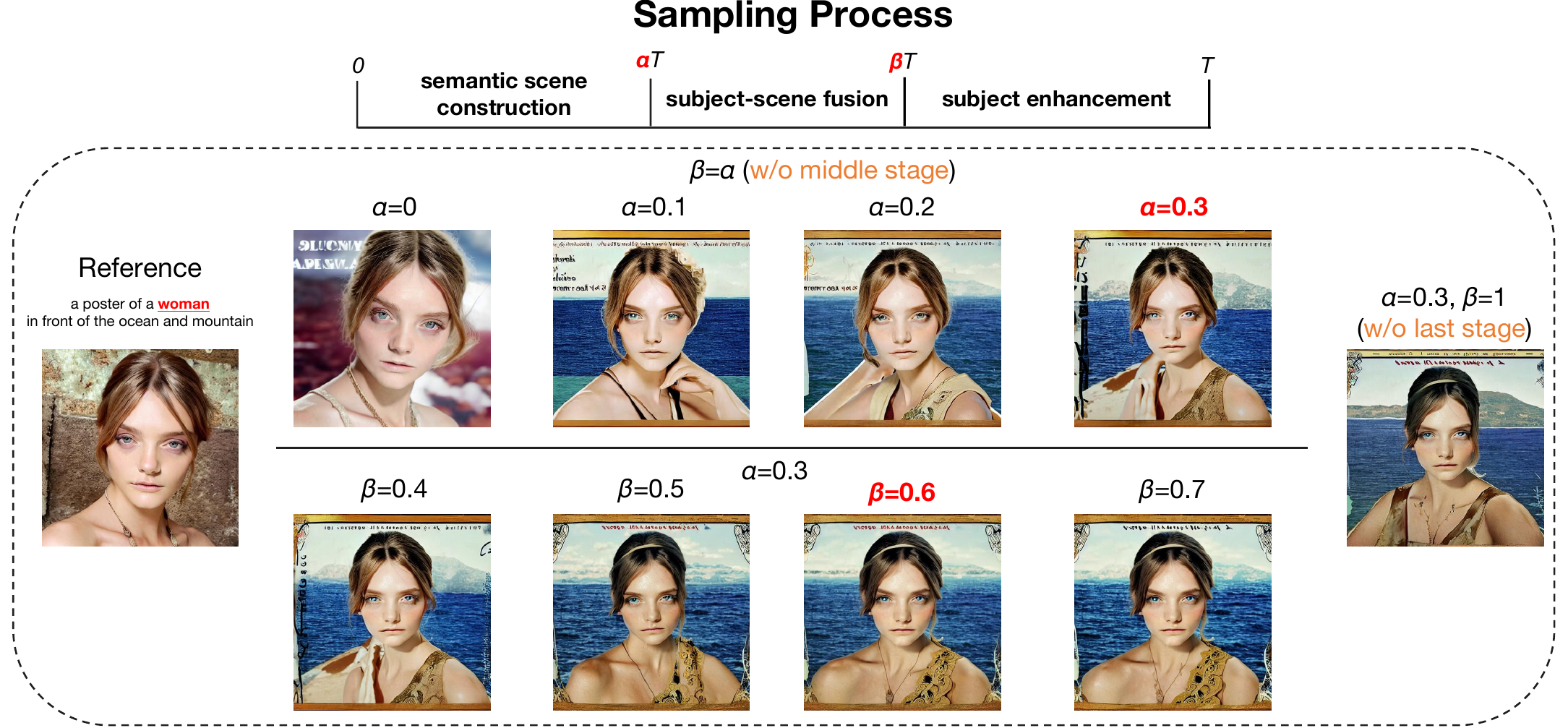}
    \caption{More hyper-parameter visualized analysis of $\alpha$ and $\beta$.}
    \label{fig:effectiveness_supply}
\end{figure*}

Additional hyper-parameter analyses are presented in \cref{fig:effectiveness_supply}.

\section{More visualized salience maps}
\begin{figure*}[]
    \centering
    \includegraphics[width=\linewidth]{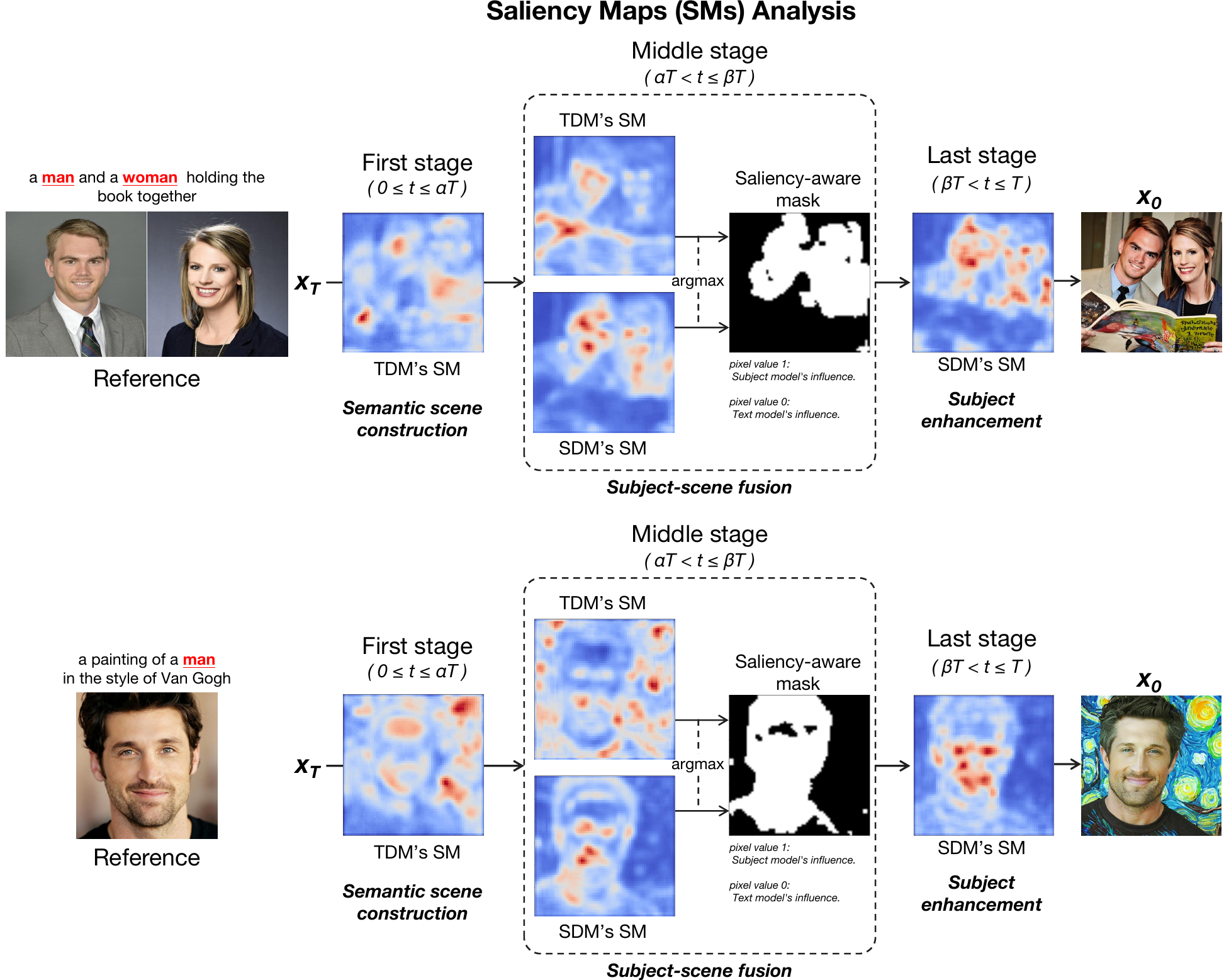}
    \caption{More cases of visualized salience maps of pre-trained
models in each stage.}
    \label{fig:visualize_supply}
\end{figure*}

Additional visualized salience maps are presented in \cref{fig:visualize_supply}.

\section{Limitation}
\label{sec:limit}
First, the persons generated by Face-diffuser closely match the reference images, which may inadvertently contribute to privacy and security concerns. It may cause 
the unauthorized use of face portraits, impacting the widespread adoption and ethical considerations. Additionally, our approach encounters challenges when it comes to editing attributes of given persons. Moving forward, we plan to engage in further research aimed at addressing these limitations and expanding the capabilities of our model.

\section{Societal impact}
The societal impact of subject-driven text-to-image generation technologies, such as Face-diffuser, is noteworthy. These advancements have far-reaching implications, fueling creativity in entertainment, virtual reality, and augmented reality industries. They enable more realistic content creation in video games and films, enhancing the overall user experience. However, as these technologies become more accessible, concerns about privacy, consent, and potential misuse have surfaced. Striking a balance between innovation and ethical considerations is crucial to harnessing the full potential of subject-driven text-to-image generation for the benefit of society.


\end{document}